\documentclass[twoside,11pt]{article}

\usepackage{blindtext}

\usepackage{jmlr2e}

\usepackage[utf8]{inputenc} 
\usepackage{hyperref}       
\hypersetup{hidelinks}
\usepackage{url}        
\usepackage{booktabs}       
\usepackage{amsfonts}       
\usepackage{nicefrac}       
\usepackage{microtype}      
\usepackage{graphicx}
\usepackage{doi}
\usepackage{rotating}
\usepackage{algorithm}
\usepackage{algpseudocode}
\usepackage{amsmath}
\usepackage{subcaption}
\usepackage{tabularx}

\usepackage{mathtools,upgreek,eucal,mathrsfs,enumitem,amsmath}
\usepackage{makeidx,dcolumn,multirow,colortbl,bm}
\usepackage[flushleft]{threeparttable}

\DeclareMathOperator{\logit}{logit}
\DeclareMathOperator{\expit}{expit}
\DeclareMathOperator{\var}{Var}

\newcommand{\bzeta}{\bm{\zeta}}

\newcommand{\btheta}{\bm{\theta}}

\newcommand{\bmeta}{\bm{\eta}}

\newcommand{\bX}{{\bf X}}
\newcommand{\bx}{{\bf x}}

\newcommand{\bz}{{\bf z}}
\newcommand{\bZ}{{\bf Z}}

\newcommand{\R}{{\mathbb R}}

\usepackage{lastpage}
\jmlrheading{27}{2026}{1-\pageref{LastPage}}{7/25; Revised
6/26}{9/26}{25-1665}{Zhanli Wu, Fabrizio Leisen, and F. Javier Rubio}
\ShortHeadings{Conformalized Regression for Continuous Bounded Outcomes}{Wu, Leisen, and Rubio}

\firstpageno{1}

\begin{document}

\title{Conformalized Regression for Continuous Bounded Outcomes}

\author{\name Zhanli Wu \email zhanli.wu@kcl.ac.uk \\
       \addr Department of Mathematics\\
       King's College London\\
       London, UK
       \AND
       \name Fabrizio Leisen \email fabrizio.leisen@kcl.ac.uk\\
       \addr Department of Mathematics\\
       King's College London\\
       London, UK
       \AND
       \name F. Javier Rubio \email f.j.rubio@ucl.ac.uk \\
       \addr Department of Statistical Science\\
       University College London\\
       London, UK
       }

\editor{Russell Greiner}

\maketitle

\begin{abstract}

Regression problems with continuous bounded outcomes frequently arise in statistical and machine learning applications, such as the analysis of rates and proportions. A central challenge in this setting is predicting the response at a new covariate value. Most of the existing literature has focused either on point prediction or on interval prediction based on asymptotic approximations.
We develop conformal prediction intervals for bounded outcomes within the framework of transformation regression models, encompassing widely used models such as beta regression and logit-normal regression. We construct non-conformity scores based on model-aligned residuals and identify a quantile-residual score that is particularly well suited to bounded outcomes, bridging normalized conformal prediction and distributional conformal prediction. This score accounts for both the heteroscedasticity inherent in such data and the asymmetry that emerges near the boundaries of the response space.
We establish marginal validity and asymptotic conditional validity for both full and split conformal prediction, holding under model misspecification. A comprehensive simulation study confirms that both methods empirically attain valid finite-sample coverage, including cases under model misspecification. A real-data application demonstrates their practical performance against bootstrap-based alternatives.

\end{abstract}

\begin{keywords}
  Beta regression, Conformal prediction, Logit-Normal, Non-conformity measure, Residuals
\end{keywords}

\section{Introduction}\label{sec:intro}

In many practical applications, the outcomes of interest take values on a bounded interval $(a, b)$. Examples include rates or proportions of events \citep{ferrari:2004,geissinger:2022}, as well as loss given default, that is, the proportion lost when borrowers default \citep{tomarchio:2019}.
A common approach to modelling such outcomes with covariates is to first rescale them to the interval $(0,1)$ and then map them to the real line through a function such as the $\logit$ \citep{smithson:2006}.
Once the transformed outcomes take values on the real line, they can be modelled using methods from generalized linear models, generalized additive models, distributional regression, or machine learning techniques such as random forests or neural networks.
In contrast, beta regression models the outcome directly as beta-distributed on the interval $(0,1)$, incorporating covariates through link functions applied to the mean parameter $\mu$ and, optionally, the precision parameter $\phi$, thereby accounting for the bounded and heteroscedastic nature of the data. This is in line with distributional regression \citep{kneib:2021}, where the aim is to model the parameters of a baseline distribution using linear or additive predictors. Since its introduction in the early 2000s \citep{kieschnick:2003,ferrari:2004}, beta regression has received considerable attention, leading to efficient software implementations \citep{cribari:2010,grun:2012}, extensions to machine learning contexts \citep{schmid:2013,weinhold:2020}, and inflated models that accommodate the presence of zeros and/or ones \citep{ospina:2012,tomarchio:2019}. Other approaches for modelling continuous bounded outcomes include Kumaraswamy regression \citep{mitnik:2013,pumi:2020} and continuous binomial regression models \citep{lee:2025}.

Beyond analyzing the structural and inferential properties of transformation or beta regression models, a question of practical interest is how to construct point and interval predictions for new covariate values \citep{ogundimu:2018}, a task that, despite its relevance, has received comparatively little attention. To the best of our knowledge, prediction for bounded outcomes has relied primarily on traditional inferential approaches, including bootstrap methods \citep{espinheira:2014}. These methods are typically grounded in asymptotic theory and may require large sample sizes to ensure accurate coverage.
Within this prediction framework, conformal prediction \citep{shafer:2008,papadopoulos:2008,vovk:2022} offers a machine learning approach for constructing prediction intervals with finite-sample marginal validity and asymptotic conditional validity. Its core component is a non-conformity measure, a function that quantifies how unusual, or non-conforming, a data point is relative to a model or algorithm trained on the remaining data. The main assumption underlying conformal prediction is exchangeability of the observations, which is often reasonable and allows the method to be coupled with a wide range of statistical models and machine learning algorithms. This assumption can, however, be violated in many regression settings.
Extensions of conformal prediction to general regression models have been studied extensively \citep{tibshirani:2019}, together with several proposals for constructing non-conformity measures in this context \citep{kato:2023}. A particular challenge in regression is heteroscedasticity, that is, when the variance (or dispersion) of the response changes with the covariates. In beta regression, the mean and variance are intrinsically linked, so that modelling the mean inherently induces heteroscedasticity. Addressing heteroscedasticity in conformal prediction is itself challenging, since its nature varies across models and outcome types. Several adaptations have been proposed to account for it, including conformalized quantile regression \citep{romano:2019}, which combines conformal prediction with quantile regression, as well as normalized conformal prediction \citep{papadopoulos:2008,lei:2018,kato:2023} and Mondrian conformal prediction \citep{dewolf:2025}, which both account for heteroscedasticity through the non-conformity measure.
A model-based extension of conformal prediction, tailored to distributional regression, was proposed by \cite{chernozhukov:2021}. Their approach uses the probability integral transform to define non-conformity measures based on an estimated conditional distribution of the response, leading to valid conformal prediction intervals in heteroscedastic settings under some regularity conditions. 

Despite the generality of existing conformal prediction frameworks, bounded outcomes present specific challenges that are not adequately addressed by standard theory. First, non-conformity scores based on raw or normalized residuals, standard in heteroscedastic regression \citep{lei:2018, kato:2023}, produce symmetric prediction intervals that can extend outside the natural support $(0,1)$; truncating to $(0,1)$ restores validity trivially but produces uninformative intervals near the boundaries. Second, heteroscedasticity in bounded outcome regression is not merely an additional modelling assumption but a structural consequence of the support: since $\text{Var}(Y) \leq \mu(1-\mu)$ for any $Y \in (0,1)$, where $\mu = {\mathbb E}(Y)$, the variance is constrained to vanish near the boundaries, inducing systematic skewness in the score distribution that generic normalization strategies do not account for. 
Although the distributional conformal prediction framework of \citet{chernozhukov:2021} is stated in sufficient generality to accommodate bounded outcomes in principle, it remains generic and does not instantiate regression models tailored to outcomes on $(0,1)$, which requires specialization to a model class with support-constrained heteroscedasticity and verification of the corresponding regularity conditions.

In this work, we propose the use of conformal prediction for regression models with continuous bounded outcomes within the framework of transformation models. This framework includes models of practical interest such as beta regression models and logit-normal regression, which are obtained for different types of transformations of the bounded outcomes.
To address heteroscedasticity in both classes of models, we construct non-conformity scores based on model-derived residuals \citep{dunn:1996}. For comparison, we also consider raw and Pearson residuals, as these are standard choices in practice. This formulation allows for bridging normalized conformal prediction and distributional conformal prediction.
Our main contribution is the identification of a quantile-residual-based score that is particularly well suited to bounded outcomes. This score simultaneously accommodates boundedness, heteroscedasticity, and the asymmetry induced near the boundaries of the response space. The novelty of our approach therefore lies not in the residual definitions themselves, but in their principled use for constructing conformal prediction intervals, together with the corresponding validity guarantees. Our development proceeds at three levels of generality. First, we present the split and full conformal prediction algorithms in full generality, so that they apply to any suitable model coupled with a suitable non-conformity score. Second, the theory narrows this scope: we characterize the conditions on the underlying model and score under which the resulting intervals attain finite-sample marginal validity and asymptotic conditional validity. Third, we show that beta regression and logit-normal regression models, coupled with the residual-based scores we propose, satisfy these conditions, so that the general guarantees apply concretely to both classes.

The remainder of the paper is organized as follows. Section~\ref{sec:regmods} describes the modelling approach based within the framework of transformation parametric regression models, and details the main models of interest in this work: beta regression and logit-normal regression. Section~\ref{sec:CPbetatreg} introduces the non-conformity measures tailored to these models and presents the split and full conformal prediction algorithms in their general form. Section~\ref{sec:theory} establishes the conditions for marginal validity and asymptotic conditional validity, and verifies them for the two model classes. Section~\ref{sec:simulation} reports an extensive simulation study of the coverage properties of split and full conformal prediction. Section~\ref{sec:applications} presents a real-data application and compares conformal prediction with bootstrap-based prediction. Finally, Section~\ref{sec:discussion} concludes with a general discussion and possible extensions. R code and data are available in the GitHub repository \url{https://github.com/ZWU-001/CPBounded}.

\section{Regression Models for Continuous Bounded Outcomes}\label{sec:regmods}

In following sections, we focus, without loss of generality, on observations indexed by $i=1,\dots,n$. For each observation, let $W_i\in (0,1)$ denote the response random variables, potentially after a linear transformation from a bounded interval $(a,b)$, where $-\infty < a < b < \infty$. We write $w_i$ for a realization of $W_i$. Let $\bX_i \in \R^p$ denote the corresponding covariate vector, and let $\bx_i$ be its observed value. To describe the different modelling approaches within a common framework, let $T:(0,1)\to\mathcal Y\subseteq\mathbb R$ be a transformation mapping the bounded response space to a subset of the real line. We define the transformed-scale response by $Y_i=T(W_i)$ and $y_i=T(w_i)$. There are two cases of practical interest:
\begin{itemize}
    \item[(i)] the case where $T(w) = w$, the identity function, and a parametric model is assumed for such bounded outcomes directly, and
    \item[(ii)] the case where $T:(0,1)\to {\mathbb R}$ is a continuous map, which allows for using any parametric regression model for unbounded outcomes. 
\end{itemize}
We consider the use of parametric regression models $Y_i \mid \bX_i = \bx_i \sim F(\cdot ; \bzeta, \bx_i)$, where $F$ is a cumulative distribution function (CDF) supported on the corresponding working scale, and $\bzeta$ denotes the model parameter vector. The conformal prediction algorithms for the new response $Y_{\text{new}}$ given a new covariate $\bX_{\text{new}}$ in Section \ref{sec:CPbetatreg} will be presented for general choices of $F$, while Section \ref{sec:theory} will characterize the conditions of such models under which theoretical guarantees are obtained. In case (i), the regression model is specified directly on the original bounded response scale. The main example considered here is beta regression, described in Section \ref{subsec:betareg}; other bounded-response models, such as Kumaraswamy regression and continuous binomial regression, could also be incorporated within the same framework.
In case (ii), the response is firstly mapped to an unbounded working scale, allowing the use of parametric, semiparametric, or nonparametric methods, such as a linear predictor, splines, Gaussian processes, or random forests, to name but a few. In this work, we focus on logit-normal regression models with possible heteroscedasticity, as described in Section \ref{subsec:treg}. These two models are presented for clarity and practical relevance, but other transformations and regression models can be incorporated into the same framework.

\subsection{Beta Regression}\label{subsec:betareg}

We now describe beta regression, which corresponds to case (i) and is of particular interest in practice for modelling continuous responses on the unit interval.
For clarity, let $Y\in (0,1)$ be a random variable with beta distribution and shape parameters $\alpha>0$, and $\beta>0$. The probability density function (PDF) of $Y$, evaluated at $y\in(0,1)$, is

\begin{equation*}
    f_{\text{Beta}}(y; \alpha, \beta) = \frac{\Gamma(\alpha + \beta)}{\Gamma(\alpha)\Gamma(\beta)} y^{\alpha - 1} (1 - y)^{\beta - 1}.
\end{equation*}

\noindent The parameters $(\alpha, \beta)$ can be re-parametrized into $(\mu,\phi)$, where $\mu = \frac{\alpha}{\alpha+\beta}\in (0,1)$ is the mean and $\phi = \alpha+\beta > 0$ represents the precision or dispersion parameter \citep{cribari:2010}. The corresponding PDF is \citep{cribari:2010}

\begin{equation*}
    f_{\text{Beta}}(y; \mu, \phi) = \frac{\Gamma(\phi)}{\Gamma(\mu\phi)\Gamma((1 - \mu)\phi)} y^{\mu\phi - 1} (1 - y)^{(1 - \mu)\phi - 1}.
\end{equation*}
Under this parametrization, $E(Y) = \mu$, and $ \var (Y)=\frac{\mu(1-\mu)}{1+\phi}$. We denote the corresponding beta cumulative distribution function (CDF) by $F_{\text{Beta}}(y;\mu,\phi)$. Beta regression (to the mean) is formulated by modelling the mean of each individual, appropriately transformed into $(0,1)$, using a linear predictor \citep{kieschnick:2003,ferrari:2004}. Specifically, the model takes the form $\varphi(\mu_i) = \theta_0 + \bx_i^{\top}\btheta$, where $\theta_0$ is the intercept,  $\btheta\in \R^p$ are the regression coefficients, and $\varphi:(0,1)\to \R$ is a continuous monotonic link function (typically, the $\logit$ function). A natural consequence of modelling the mean in beta regression is that it implicitly defines a model for the variance as well. As a result, beta regression is inherently heteroscedastic.
The maximum likelihood estimates (MLEs) of ${\theta_0}$, ${\btheta}$ and $\phi$ can be obtained by maximizing the corresponding likelihood function. The MLEs are not available in closed-form, however, efficient numerical methods and implementations are available in R package \texttt{betareg} and other programming languages \citep{cribari:2010}. Once the MLEs $\widehat{\theta_0}$, $\widehat{\btheta}$ and $\widehat{\phi}$ have been obtained, predictions for a new covariate value $\bx_{\text{new}}$, can be obtained as the predicted mean response $\widehat{y}_{\text{new}} = \widehat{\mu}_{\text{new}} = \expit(\widehat{\theta_0} + \bx_{\text{new}}^{\top}\widehat{\btheta})$, where $\expit(\cdot)$ denotes the inverse of the $\logit$ function.
A natural extension of the beta regression model to the mean consists of modelling $\phi$ using covariates. Typically, covariates are included in this parameter through the log-linear model $\phi_i = \exp\left\{ \eta_0 + \bx_i^{\top}\bmeta \right\}$, where $\eta_0$ is the intercept and $\bmeta \in \R^p$ are the regression coefficients.

\subsection{Logit-Normal Regression}\label{subsec:treg}

We now consider case (ii), focusing on logit-normal regression as a particular example of a transformation regression model. 
Transformation regression models are obtained by mapping the values $w_i \in (0,1)$, onto the real line through a continuous, strictly increasing, bijection $T:(0,1)\to\R$. We consider models of the form
\begin{equation*}
    y_i=T(w_i) = \psi(\bx_{i}) + \epsilon_{i},
\end{equation*}
where $\mu_i=\psi(\bx_i) = E[T(w_i)\mid \bx_i]$ is the regression function on the transformed scale and the errors satisfy $E[\epsilon_i \mid \bx_i] = 0$. In this work, we focus on the parametric specification $\psi(\bx_{i}) = \theta_0 + \bx_{i}^{\top}\btheta$ and $\epsilon_{i} \stackrel{ind.}{\sim} {N}(0, \sigma_i^2)$, where $\theta_0$ is the intercept and $\btheta \in \R^p$ are the regression coefficients in the model for the conditional mean. To allow for heteroscedasticity, the conditional standard deviation is modelled through a separate log-linear predictor, $\sigma_i = \exp\{\eta_0 + \bx_{i}^{\top}\bmeta\}$, where $\eta_0$ and $\bmeta$ are corresponding intercept and regression coefficients. The homoscedastic case is recovered by setting $\bmeta=\bm{0}$, so that $\sigma_1 = \dots = \sigma_n = \sigma$. An implementation of this model can be found in the R package \texttt{gamlss} \citep{stasinopoulos:2008}. 


Of particular interest is the case where $T(w) = \logit(w) = \log\left(\dfrac{w}{1-w}\right)$, which leads to classical logit-normal regression. That is, this formulation corresponds to a linear regression model on the logit scale. Consequently, this strategy can be coupled with any tool available for classical linear regression. 
After fitting this model, the predicted values for a new covariate value $\bx_{\text{new}}$ can be obtained as $\widehat{y}_{\text{new}} = \widehat{\mu}_{\text{new}}=\widehat{\theta}_0+\bx_{\text{new}}^{\top}\widehat{\btheta}$ and accordingly, $\widehat{w}_{\text{new}} = \expit(\widehat{y}_{\text{new}})=\expit(\widehat{\theta}_0+\bx_{\text{new}}^{\top}\widehat{\btheta})$, where $\widehat{\theta}_0$ and $\widehat{\btheta}$ denote the estimated intercept and regression coefficient vector.

\section{Conformal Prediction for Transformation and Regression Models}\label{sec:CPbetatreg}

In this section, we present the proposed strategy for constructing conformal prediction intervals for continuous bounded responses using the regression models introduced in Section \ref{sec:regmods}. Throughout this section, let $\mathcal{D}_n=\{(w_i,\bx_i)\}_{i=1}^n$ denote the observed data, where $w_i\in(0,1)$, and let $y_i=T(w_i)$ be the corresponding response on the working scale. We denote $W_{\text{new}}\in(0,1)$ as the new bounded response associated with the covariate vector $\bX_{\text{new}}=\bx_{\text{new}}$, and let $Y_{\text{new}}=T(W_{\text{new}})$ and $y_{\text{new}}$ be the corresponding response on the working scale and its realization. Let $F(\cdot; \widehat{\bzeta},\bx_i)$ denote the fitted CDF of the working-scale response $Y_i\mid\bX_i=\bx_i$, where $\widehat{\bzeta}$ denotes the estimated parameters based on the $\mathcal{D}_n$. For both regression models introduced in Section \ref{sec:regmods}, 
$\widehat{\bzeta}=(\widehat{\theta}_0,\widehat{\btheta},
\widehat{\eta}_0,\widehat{\bmeta})$ when heteroscedasticity is considered.
When $T$ is the identity map, so that $y_i=w_i$, this is the fitted CDF of the original bounded response, as in beta regression. When $T$ maps $(0,1)$ to the real line $\mathbb R$, this is the fitted CDF of the transformed response, as in logit-normal regression. The conformal construction is carried out on the working scale; the resulting interval is then mapped back to the original bounded response scale whenever $T$ is not the identity. For notational convenience, we use the working-scale pairs $\{(y_i,\bx_i)\}_{i=1}^n$ in the remainder of this section. We begin by introducing residual-based non-conformity measures, which are derived from model-specific residuals in general regression settings \citep{dunn:1996}. These measures are tailored to each model and account for heteroscedasticity where appropriate. We then describe the general split conformal and full conformal algorithms and focus on the models and the non-conformity measures presented in Section \ref{subsec:non-conformity}. Without loss of generality, we set $\widehat{y}_1, \dots, \widehat{y}_n$ to be the point prediction for working-scale observations $1, \dots, n$ generated by the fitted regression model. Specifically, for the beta regression model, we have $\widehat{y}_i=\expit(\widehat{\theta}_0+\bx_{i}^{\top}\widehat{\btheta})$, where $\widehat{\theta}_0$ and $\widehat{\btheta}$ denote the estimated intercept and regression coefficient vector, respectively, obtained from the fitted regression model. For the logit-normal regression model, we have $\widehat{y}_i=\widehat{\theta}_0+\bx_{i}^{\top}\widehat{\btheta}$, and accordingly, the prediction in the original scale would be $\widehat{w}_i=\expit(\widehat{\theta}_0+\bx_{i}^{\top}\widehat{\btheta})$.


\subsection{Non-Conformity Measures}\label{subsec:non-conformity}

The three scores introduced below are presented in an increased order of complexity in terms of how they incorporate information about the fitted model's structure. Raw residuals~\eqref{eq:raw_res} target only the conditional location; Pearson residuals~\eqref{eq:pearson_res} additionally standardize by the conditional scale and thus adapt to heteroscedasticity; and quantile residuals~\eqref{eq:res_quantile} use the entire fitted conditional distribution through the probability integral transform, thus also accommodating asymmetry and tail behaviour.

\begin{itemize}
\item \textbf{Raw residuals:} A raw-residuals based non-conformity measure is defined as:
\begin{equation}\label{eq:raw_res}
r_i = \vert y_i - \widehat{y}_i\vert.
\end{equation}

\item \textbf{Pearson residuals:} A Pearson-residuals based non-conformity measure is defined as:
\begin{equation}\label{eq:pearson_res}
r^{P}_{i} = \frac{|y_i - \widehat{y}_i|}{\sqrt{\widehat{\text{Var}}(Y_i \mid \bx_i)}},
\end{equation}
where for beta regression, $\widehat{\text{Var}}(Y_i \mid \bx_i) = \frac{\widehat{\mu}_i(1-\widehat{\mu}_i)}{1+\widehat{\phi}_i}$; $\widehat{\mu}_i=\widehat y_i$ is the fitted mean and $\widehat{\phi}_i$ is the fitted precision parameter; for the logit-normal regression, $\widehat{\operatorname{Var}}(Y_i\mid \bx_i)=\widehat{\sigma}_i^2$, where $\widehat{\sigma}_i$ is the fitted conditional standard deviation.
This type of residual adjusts the difference between the observed value and the fitted mean prediction by the estimated standard deviation, assuming it exists, thus accounting for heteroscedasticity through localized weighting. The resulting non-conformity measure resembles those used in normalized conformal prediction, where absolute raw residuals are scaled by general functions $\tau_i(\bx_i)$. Various choices for $\tau_i$ have been explored in the context of regression models \citep{kato:2023,dewolf:2025}, with the goal of capturing different forms of heteroscedasticity.

\item \textbf{Quantile residuals:} A quantile-residuals based non-conformity measure is defined as:
\begin{equation}\label{eq:res_quantile}
    r_i^{Q} = \left \vert \Phi^{-1} \{ F(y_i; \widehat{\bzeta}, \bx_i) \} \right \vert.
\end{equation} 
The choice of $\Phi^{-1}$ is convenient because it makes the quantile residual reduce to the standardized (Pearson) residual in the special case of a heteroscedastic normal model: there, the probability integral transform gives $F(y_i;\widehat{\bzeta}_i,\bx_i)=\Phi\{(y_i-\widehat{y}_i)/\widehat{\sigma}_i\}$, so that Equations~\eqref{eq:pearson_res} and~\eqref{eq:res_quantile} coincide, that is, $r^{P}_{i}=r^{Q}_{i}=|y_i-\widehat{y}_i|/\widehat{\sigma}_i$. This identity is specific to the Gaussian case; for non-Gaussian models such as beta regression the two scores differ, with the quantile residual additionally capturing the asymmetry and boundary behaviour of the conditional distribution.
\end{itemize}

These scores form a coherent family rather than an generic list. This is illustrated by the heteroscedastic normal case, in which the Pearson and quantile scores coincide: the quantile residual generalizes the Pearson residual to non-Gaussian, bounded conditional distributions, and recovers the distributional conformal prediction score of \citet{chernozhukov:2021} as an interpretable special case (using the uniform distribution instead of Gaussian). In this sense, the non-conformity measure is not chosen generically but aligned with the model used to fit the data, a residual-based framework that links the conformal score to the model's distributional structure and bridges normalized and distributional conformal prediction.

The interplay between heteroscedasticity and the choice of non-conformity score is central here. In transformation regression, heteroscedasticity is introduced explicitly through a covariate-dependent scale $\sigma_i$ (Section~\ref{subsec:treg}), whereas in beta regression it is intrinsic, since modelling the mean $\mu_i$ already shapes the variance, with further flexibility when $\phi_i$ is covariate-dependent (Section~\ref{subsec:betareg}). For bounded outcomes, however, the difficulty is not only one of changing variance: near the boundaries of $(0,1)$ the conditional distribution becomes asymmetric and heavier-tailed, so the symmetric Pearson-type score~\eqref{eq:pearson_res} may reflect boundary-induced skewness rather than local scale. Quantile residuals~\eqref{eq:res_quantile} instead adapt to the full conditional distribution, accounting for scale, asymmetry, and tail behaviour in a single principled measure, which is why we recommend them for (transformed) bounded outcomes.

Our framework is organized by a single principle: the non-conformity score is aligned with the residual structure of the fitted model. The quantile-residual score~\eqref{eq:res_quantile} is the member of this family obtained from the probability integral transform, which is also why it relates to the baseline distributional conformal score of \citet{chernozhukov:2021}. This coincidence, however, is a property of that particular residual and not of the framework: other residuals for beta regression \citep{dunn:1996,espinheira:2008,espinheira:2017} lead to non-conformity measures that are valid within our construction yet do not reduce to a rank-based distributional conformal score. The two viewpoints therefore intersect at the quantile residual rather than one containing the other. Among these residuals, the members differ in how directly they accommodate heteroscedasticity; for bounded outcomes we recommend the quantile residual because it adapts to the full conditional distribution, capturing scale, asymmetry, and boundary behaviour in a single measure, whereas several alternatives capture only some of these. Normalized non-conformity measures developed for heteroscedastic regression \citep{kato:2023} can also be embedded in the framework, though they are not generally tailored to the support constraint of bounded outcomes.

A related perspective is Bayesian conformal prediction, which combines posterior-predictive distributions with conformal calibration \citep{fong:2021,bersson:2024,deliu:2026}. Our quantile-residual approach is conceptually related in that it also constructs the non-conformity score from a fitted model, although here this distribution is obtained from a frequentist regression model rather than a Bayesian posterior-predictive distribution.

We note that these quantile residuals \citep{dunn:1996} are distinct from the quantile scores used in conformalized quantile regression \citep{romano:2019}. The latter are built from conditional quantile functions of the response, typically estimated by quantile regression, and measure the signed distance of the response from an estimated quantile band. By contrast, the score in \eqref{eq:res_quantile} is obtained from the fitted conditional distribution $F(\cdot; \widehat{\bzeta}, \bx_i)$ through the probability integral transform, so that conditional quantiles enter only at interval construction, when $F$ is inverted; in this sense the score is closer in spirit to the distributional conformal prediction framework \citep{chernozhukov:2021} rather than the quantile-regression family. Both approaches address heteroscedasticity, but through different mechanisms: covariate-dependent quantile bands in the former, and the covariate-dependent conditional distribution in the latter, which for bounded outcomes additionally captures boundary-induced asymmetry.

\subsection{Split and Full Conformal Prediction Algorithms}

We now present split and full conformal prediction algorithms for continuous bounded outcomes. The algorithms are stated in relatively general terms. These algorithms will later be used within the two modelling frameworks described in Sections \ref{subsec:betareg}--\ref{subsec:treg} (beta regression and logit-normal regression), together with the residual-based scores in Section \ref{subsec:non-conformity}, with particular emphasis on the quantile-residual score. 
The corresponding model-specific interval constructions are detailed in Sections \ref{subsec:split_construction} and \ref{subsec:full_construction}.

\begin{algorithm}[ht]
\caption{Split Conformal Prediction for Continuous Bounded Outcomes}
\label{alg:splitCP}
\begin{algorithmic}[1]
\Require Dataset $\{(w_i,\bx_i)\}_{i=1}^n$ with $w_i\in(0,1)$, new covariate $\bx_{\text{new}}$, significance level $\alpha$, non-conformity score $S$, continuous mapping $T:(0,1)\to\mathcal Y\subseteq\mathbb R$ with inverse $T^{-1}$.
\Ensure Split prediction set $\mathcal{C}^{\text{split}}_{1-\alpha}(\bx_{\text{new}})\subseteq(0,1)$ for the new response $y_{\text{new}}$ at $\bx_{\text{new}}$.

    \State Map the responses to the working scale: $y_i=T(w_i)$, $i=1,\ldots,n$.
    
    \State Randomly split the dataset $\{(y_i,\bx_i)\}_{i=1}^n$ into disjoint subsets $\mathcal{I}_{\text{train}}$ and $\mathcal{I}_{\text{cal}}$.
    
    \State Fit a regression model on $\{(y_i,\bx_i):i\in\mathcal{I}_{\text{train}}\}$, producing the fitted model $\widehat{\mathcal M}(\widehat{\bzeta})$.
    
    \State Compute calibration non-conformity scores
    \[
    s_i = S(y_i,\bx_i;\widehat{\mathcal M}(\widehat{\bzeta})), 
    \qquad i\in\mathcal{I}_{\text{cal}}.
    \]
    
    \State Let $q_{1-\alpha}$ be the $\left\lceil (1-\alpha)(\#\mathcal{I}_{\text{cal}}+1)\right\rceil$-th smallest value among $\{s_i:i\in\mathcal{I}_{\text{cal}}\}$, where $\#\cdot$ denotes the set size.



    \State Construct the interval on the working scale:
    \[
    \widetilde{\mathcal C}^{\text{split}}_{1-\alpha}(\bx_{\text{new}})
    =
    (\widetilde{L}_{\text{split}},\widetilde{U}_{\text{split}})
    =\left\{
    y :
    s_{\mathrm{new}}
    =
    S\!\left(
    y,\mathbf{x}_{\mathrm{new}};
    \widehat{\mathcal{M}}(\widehat{\boldsymbol{\zeta}})
    \right)
    \le q_{1-\alpha}
    \right\}.
    \]
    
    \State Map $\widetilde{\mathcal C}^{\text{split}}_{1-\alpha}(\bx_{\text{new}})$ back to $(0,1)$ by $T^{-1}$ to obtain $\mathcal C^{\text{split}}_{1-\alpha}(\bx_{\text{new}})=(L_{\text{split}},U_{\text{split}})$.

\end{algorithmic}
\end{algorithm}

\begin{algorithm}[ht]
\caption{Full Conformal Prediction for Continuous Bounded Outcomes}
\label{alg:fullCP}
\begin{algorithmic}[1]
\Require Dataset $\{(w_i,\bx_i)\}_{i=1}^n$ with $w_i\in(0,1)$, new covariate $\bx_{\text{new}}$, significance level $\alpha$, non-conformity score $S$, continuous mapping $T:(0,1)\to\mathcal Y\subseteq\mathbb R$ with inverse $T^{-1}$, candidate grid $\mathcal G\subseteq\mathcal Y$, and tolerance $\varepsilon>0$.
\Ensure Full prediction set $\mathcal{C}^{\text{full}}_{1-\alpha}(\bx_{\text{new}})\subseteq(0,1)$ for the new response $y_{\text{new}}$ at $\bx_{\text{new}}$.

    \State Map the responses to the working scale: $y_i=T(w_i)$, $i=1,\ldots,n$.

    \State Initialize an empty set $\mathcal S=\emptyset$ to store conformal candidates on the working scale.

    \For{each $y\in\mathcal G$}
        \State Augment the working-scale dataset $\{(y_i,\bx_i)\}_{i=1}^n$ with the candidate point $(y,\bx_{\text{new}})$.
        
        \State Fit a regression model on the augmented dataset, producing the fitted model $\widehat{\mathcal M}^{(y)}(\widehat{\bzeta})$.
        
        \State Compute the non-conformity scores
        \[
        s_i^{(y)}
        =
        S(y_i,\bx_i;\widehat{\mathcal M}^{(y)}(\widehat{\bzeta})),
        \qquad i=1,\ldots,n,
        \]
        and
        \[
        s_{\text{new}}^{(y)}
        =
        S(y,\bx_{\text{new}};\widehat{\mathcal M}^{(y)}(\widehat{\bzeta})).
        \]
        
        \State Let $q_{1-\alpha}^{(y)}$ be the $\left\lceil(1-\alpha)(n+1)\right\rceil$-th smallest value among 
        \[
        \{s_1^{(y)},\ldots,s_n^{(y)}\}.
        \]
        
        \If{$s_{\text{new}}^{(y)}\le q_{1-\alpha}^{(y)}$}
            \State Include $y$ in $\mathcal S$.
        \EndIf
    \EndFor



    \State Let $\widetilde{\mathcal C}^{\text{full}}_{1-\alpha}(\bx_{\text{new}})=(\widetilde L_{\text{full}},\widetilde U_{\text{full}})$ denote the lower and upper bounds of $\mathcal S$, approximated to within tolerance $\varepsilon$.

    \State Map $\widetilde{\mathcal C}^{\text{full}}_{1-\alpha}(\bx_{\text{new}})$ back to $(0,1)$ by $T^{-1}$ to obtain $\mathcal C^{\text{full}}_{1-\alpha}(\bx_{\text{new}})=(L_{\text{full}},U_{\text{full}})$.
\end{algorithmic}
\end{algorithm}

The algorithms are stated in general terms through a regression model, the non-conformity score $S$, and the fixed working-scale mapping $T$ used before model fitting. In our framework, as is discussed earlier, for beta regression no mapping is applied, so \(y=T(w)=w\), \(\mathcal Y=(0,1)\), and the model is fitted directly to the bounded response. For logit-normal regression, \(y=T(w)=\logit(w)\), \(\mathcal Y=\mathbb R\), \(T^{-1}\) is the $\expit$ map, and the model is fitted to the transformed responses. In both cases, the conformal construction is carried out on the working scale, while the final prediction interval is reported on the original bounded response scale. For clarity, we write $S(y_i,\bx_i;\widehat{\mathcal M}(\widehat{\bzeta}))$ in Algorithm \ref{alg:splitCP} for the non-conformity score assigned to a candidate pair $(y_i,\bx_i)$, where $y_i \in\mathcal Y$ is a response value on the working scale, $\bx_i \in\mathbb R^p$ is the associated covariate vector, and $\widehat{\mathcal M}(\widehat{\bzeta})$ denotes the fitted regression model with estimated parameter vector $\widehat{\bzeta}$. The fitted model provides the quantities needed to evaluate the score, such as the fitted conditional location, conditional variance, or fitted conditional CDF. $\widehat{\mathcal M}^{(y)}(\widehat{\bzeta})$ in Algorithm \ref{alg:fullCP} represents the fitted model with the candidate value $y$.

In Step~7 of Algorithm~\ref{alg:splitCP}, since $T$ is strictly monotone in our modelling framework, $\mathcal C^{\text{split}}_{1-\alpha}(\bx_{\text{new}})$ is an interval \citep{chernozhukov:2021}. In Algorithm~\ref{alg:fullCP}, both the refitted model and the threshold depend on the candidate value, so $\mathcal S$ need not be connected. If $\mathcal S$ is connected, $\widetilde L^{\mathrm{full}}$ and $\widetilde U^{\mathrm{full}}$ are the endpoints of the corresponding full conformal prediction interval. If $\mathcal S$ is disconnected, $\widetilde{\mathcal C}^{\mathrm{full}}_{1-\alpha}(\bx_{\mathrm{new}})$ is the convex hull of the full conformal prediction set on the working scale. Mapping these endpoints back through the strictly monotonic transformation $T^{-1}$ gives the reported interval on the original response scale, which we denote by
$\overline{\mathcal C}^{\mathrm{full}}_{1-\alpha}(\bx_{\mathrm{new}})$.
Thus, $\overline{\mathcal C}^{\mathrm{full}}_{1-\alpha}(\bx_{\mathrm{new}})$ coincides with the exact full conformal prediction set when the latter is connected, and is its convex hull otherwise, following standard practice in full conformal prediction
\citep[][report the closed interval spanned by the accepted values]{chernozhukov:2021}. In our simulation studies and applications, we did not find cases with disconnected sets. Nevertheless, they may arise in small samples under particular data-generating mechanisms.
Since taking the convex hull can only enlarge the prediction set, the coverage guarantee is retained, at the cost of some conservativeness in the reported width when the set is disconnected, as shown in the next section. In cases where the disconnected components themselves are of interest, they can be obtained from the accepted grid values by repeating the binary search on several regions of $(0,1)$, though interpreting and calculating them requires a case-by-case inspection.

The split conformal method is computationally efficient, as it fits the regression model only once, making it well suited for large datasets. However, it relies on a random split of the data into training and calibration sets, which may reduce statistical efficiency and introduce variability in the resulting prediction intervals. In contrast, the full conformal method uses the entire dataset for both model fitting and calibration by refitting the model for each candidate response value. This typically produces narrower intervals but incurs substantially higher computational cost due to repeated model training across the prediction grid. These can be observed from empirical results in Section \ref{sec:simulation}--\ref{sec:applications}.

\subsection{Split Conformal Prediction Intervals Construction}\label{subsec:split_construction}
Algorithm \ref{alg:splitCP} is formulated generically and can be applied to a range of regression models and non-conformity score functions. In this section, we describe the corresponding model-specific constructions of the split conformal prediction intervals. In each case, the prediction interval is obtained by solving the inclusion inequality in Step~6 of Algorithm \ref{alg:splitCP}

\begin{equation}
    \left\{
    y:
    S(y,\bx_{\text{new}};\widehat{\mathcal M}(\widehat{\bzeta}))\le q_{1-\alpha}
    \right\}.
    \label{eq:scores}
\end{equation}

With the new covariate $\bx_{\text{new}}$ and the fitted model $\widehat{\mathcal M}(\widehat{\bzeta})$, we can obtain the new point prediction $\widehat{y}_{\text{new}}$. The explicit form of this inequality depends on the chosen non-conformity score. For example, the raw residual score depends only on the fitted conditional location $\widehat y_{\text{new}}$, the Pearson residual score additionally depends on the fitted conditional variance or scale at $\bx_{\text{new}}$, and the quantile residual score depends on the full fitted conditional CDF $F(\cdot;\widehat{\bzeta},\bx_{\text{new}})$.

\begin{itemize}
    \item \textbf{Beta regression model:} For this model, no transformation is applied to the response. Two variants of the beta regression model are considered: one in which only the mean parameter \( \mu \) depends on covariates, and another one where both the mean \( \mu \) and the precision parameter \( \phi \) are modelled as functions of covariates. For each scenario, two types of nonconformity scores are employed: the Pearson residual in \eqref{eq:pearson_res} and the quantile residual in \eqref{eq:res_quantile}. Under the Pearson residual, \eqref{eq:scores} becomes 
    
    $$\left\lbrace y: \frac{\left \vert y-\widehat{y}_{\text{new}} \right \vert}{\sqrt{\widehat{\text{Var}}(Y_{\text{new}} \mid \bx_{\text{new}})}} \leq q_{1-\alpha} \right\rbrace. $$ 
    
    This results in the interval
    
    $$\left(\max\left(0,\widehat{y}_{\text{new}}-q_{1-\alpha}\sqrt{\widehat{\text{Var}}(Y_\text{new} \mid \bx_\text{new})}\right),\min\left(1, \widehat{y}_{\text{new}}+q_{1-\alpha}\sqrt{\widehat{\text{Var}}(Y_\text{new} \mid \bx_\text{new})}\right)\right),$$ 
    
    where $\sqrt{\widehat{\text{Var}}(Y_\text{new} \mid \bx_\text{new})}=\sqrt{\widehat{y}_{\text{new}}(1 - \widehat{y}_{\text{new}})/(1 + \widehat{\phi}_{\text{new}})}$ and \(\widehat{\phi}_{\text{new}}\) denotes the estimated dispersion of the new observation (including both homoscedastic and heteroscedastic scenarios). In this setting, the resulting interval may extend beyond the unit interval \((0, 1)\). This limitation arises because the interval construction accounts only for heteroscedasticity through the variance term \(\sqrt{\widehat{\text{Var}}(Y_\text{new}\mid \bx_\text{new})}\), while ignoring the skewness of the beta distribution near the boundaries. When \(\widehat{y}_{\text{new}} \to 0^+\), residuals exhibit positive skewness; when \(\widehat{y}_{\text{new}} \to 1^-\), negative skewness prevails. 
    To ensure that the prediction interval respects the bounded response space, we truncate it to \((0,1)\). This truncation is not part of the theoretical construction in Section \ref{sec:theory} and does not affect the validity guarantees; since $Y_{\text{new}} \in (0,1)$ almost surely for beta models, the coverage event is unchanged. 
    
    Under the quantile residual, \eqref{eq:scores} becomes 
    
    $$\left\{y: \left \vert \Phi^{-1} \{ F_{\text{Beta}}(y; \widehat{\mu}_{\text{new}}, \widehat{\phi}_{\text{new}}) \} \right \vert \leq q_{1-\alpha}\right\}.$$ 
    
    This results in the interval
    
    $$\left(F^{-1}_{\text{Beta}}\left(\Phi(-q_{1-\alpha}); \widehat{\mu}_{\text{new}},\widehat{\phi}_{\text{new}}\right), F^{-1}_{\text{Beta}}\left(\Phi(q_{1-\alpha}); \widehat{\mu}_{\text{new}},\widehat{\phi}_{\text{new}}\right)\right),$$ 

    where \(F^{-1}_{\text{Beta}}\) is the inverse beta CDF. This inverse transformation ensures the interval remains within \((0, 1)\), naturally accounts for skewness, and eliminates the need for truncation.

    \item \textbf{Logit-Normal regression model:} Under the homoscedastic setting ($\sigma_1 = \dots = \sigma_n = \sigma$), we use the absolute raw residual from \eqref{eq:raw_res} as the score function. Here $\widehat{y}_{\text{new}}$ denotes the fitted conditional mean on the transformed scale. \eqref{eq:scores} then becomes 
    
    $$\left\{y: \left \vert y-\widehat{y}_{\text{new}} \right \vert \leq q_{1-\alpha}\right\}. $$
    
    This results in the prediction interval
    
    $$(\expit(\widehat{y}_{\text{new}}-q_{1-\alpha}),\expit(\widehat{y}_{\text{new}}+q_{1-\alpha})).$$ 
    
    The function $\expit$ is used to map the prediction interval in $\mathbb{R}$ back to the $(0,1)$ scale. Under the heteroscedastic setting ($\sigma_i$'s are covariate-dependent), we use the absolute Pearson residual from \eqref{eq:pearson_res}. In this case, inequality \eqref{eq:scores} becomes 
    
    $$\left\{y: \frac{\left \vert y-\widehat{y}_{\text{new}} \right \vert}{\widehat{\sigma}_{\text{new}}} \leq q_{1-\alpha}\right\}. $$ 
    
    This results in the interval form
    
    $$\left(\expit(\widehat{y}_{\text{new}}-q_{1-\alpha}\widehat{\sigma}_{\text{new}}), \expit(\widehat{y}_{\text{new}}+q_{1-\alpha}\widehat{\sigma}_{\text{new}})\right),$$ 
    
    where $\widehat{\sigma}_{\text{new}}$ is the estimated standard deviation of the new observation.
    
\end{itemize}

\subsection{Full Conformal Prediction Intervals Construction}\label{subsec:full_construction}

A straightforward implementation of full CP (Algorithm \ref{alg:fullCP}) evaluates the non-conformity score at every candidate response value $y$ on a predefined grid $\mathcal{Y}$, augmenting and refitting the model for each candidate, and then identifies the smallest and largest values satisfying the conformal inclusion criterion. Although conceptually simple, this exhaustive search can be computationally demanding when the required precision level is high or the fitted model is complex. We therefore employ an adaptive interval-finding algorithm that locates the conformal interval boundaries without evaluating every grid point. In the case of beta regression models, we numerically solve for the interval within the unit interval \( (0,1) \) by applying an interval-finding algorithm with a user-specified tolerance $\varepsilon>0$. Based on Algorithm \ref{alg:fullCP}, our aim is to identify the set

\begin{equation*}
    \mathcal{C}^{\text{full}}_{1-\alpha}(\bx_{\text{new}}) = \{ y \in (0,1) : S(y,\bx_{\text{new}};\widehat{\mathcal M}^{(y)}(\widehat{\bzeta})) \leq q_{1-\alpha}^{(y)} \}.
\end{equation*}

To check whether a candidate value belongs to $\mathcal{C}^{\text{full}}_{1-\alpha}$, we define an indicator-type function

\[
I: (0, 1) \rightarrow \{\texttt{TRUE}, \texttt{FALSE}\},
\]

where $I(\xi) = \texttt{TRUE}$ if the augmented nonconformity score for candidate $y$ satisfies the inclusion criterion, and \texttt{FALSE} otherwise. Our objective is to find the values $L$ and $U$, that is, lower bound and upper bound for the prediction interval, such that $I(\xi) = \texttt{TRUE}$ for all $\xi \in [L,U] \subset (0,1)$. The core idea is to locate a point inside the prediction interval, that is, satisfying the inclusion criterion. We decide an initial guess $\xi_0\in(0,1)$, which leads to two cases.

\paragraph{Case 1: The initial guess lies within the inclusion region ($I(\xi_0) = \texttt{TRUE}$).}

In this case, the algorithm proceeds as follows.

\begin{itemize}

    \item Find the lower endpoint $L$. A binary search is performed on the interval $(0, \xi_0)$, repeatedly halving subintervals to locate the point where $I(\xi)$ transitions from \texttt{FALSE} to \texttt{TRUE} within tolerance $\varepsilon$.
    
    \item Find the upper endpoint $U$. A similar binary search is performed on $(\xi_0, 1)$, locating the transition from \texttt{TRUE} to \texttt{FALSE}.
\end{itemize}

\paragraph{Case 2: The initial guess lies outside the inclusion region ($I(\xi_0) = \texttt{FALSE}$).}

In this case, we need to perform an exploratory search to find a point inside the interval.

\begin{itemize}

    \item \textbf{Initialization:} A coarse grid is set up over $(0,1)$, typically starting with $\{0,0.5,1\}$. The endpoints $0$ and $1$ are assigned with \texttt{FALSE} values.
    
    \item \textbf{Grid Expansion:} If no \texttt{TRUE} values are detected on the initial grid, that is, $I(0.5)=\texttt{FALSE}$, the grid is refined iteratively by inserting midpoints between all adjacent pairs. This process continues until a point $\xi' \in (0,1)$ is found such that $I(\xi') = \texttt{TRUE}$.
    
    \item \textbf{Boundary Localization:} Once a valid inclusion point $\xi'$ is located, its immediate \texttt{FALSE} neighbors on both sides can be identified. These provide bounds within which the transitions occur.
    
    \item \textbf{Binary Search:} The lower and upper boundaries of the inclusion set are then refined within tolerance $\varepsilon$ using the same binary search idea in Case 1.
    
\end{itemize}

For logit-normal regression models under the full conformal setting, the candidate response values need to be searched over a grid on the transformed real line. One choice to define the candidate grid $\mathcal{Y}$ is using an expanded interval around, for example, the classical prediction bounds in normal linear regression:

\[
\left[l-\rho_{\text{grid}}(u-l), u+\rho_{\text{grid}}(u-l)\right],
\]

where \( l \) and \( u \) denote the lower and upper bounds of the classical prediction interval on the transformed scale. The parameter \(\rho_{\text{grid}}\) is only a numerical grid-expansion parameter for this specific full conformal implementation. It is not a conformal tuning parameter and does not enter the theoretical validity result. Any sufficiently large choices would lead to the same conformal interval up to the grid resolution, provided that the grid contains the conformal inclusion region. In practice, this can be checked by verifying that the conformal inclusion rule is false at both endpoints; otherwise, the grid needs to be enlarged. To construct the prediction interval, similarly, we firstly set an initial guess $\xi_0\in\mathbb{R}$. If $I(\xi_0) = \texttt{TRUE}$, we perform the binary search on the interval $(l-\rho_{\text{grid}}(u-l), \xi_0)$ and the interval $(\xi_0,u+\rho_{\text{grid}}(u-l))$ similar to Case 1. If $I(\xi_0) = \texttt{FALSE}$, we adopt a similar strategy outlined in Case 2. The only difference is that instead of inserting midpoints of candidate values, we insert the values corresponding to the floor of the midpoint indices.

To reduce computational cost, evaluations of $I(\xi)$ are stored and reused (that is, memorized), thereby avoiding redundant computation during grid refinement and binary search phases.
The theoretical results in Section \ref{sec:theory} concern the exact conformal set $\mathcal{C}^{\mathrm{full}}_{1-\alpha}$. The procedure described above is a numerical implementation that approximates the endpoints of this set up to the user-specified tolerance $\varepsilon$, provided that the initial search domain contains the conformal inclusion region. Accordingly, neither $\varepsilon$ nor $\rho_{\mathrm{grid}}$ enters the theoretical validity results.

\section{Theoretical Guarantees}\label{sec:theory}

We now establish theoretical guarantees for conformal prediction intervals for bounded outcomes based on Algorithms~\ref{alg:splitCP}--\ref{alg:fullCP}. Proposition~\ref{prop:marginal} establishes finite-sample marginal validity for both split and full conformal prediction under exchangeability. Although exchangeability-based marginal validity for generic conformal prediction is well established \citep{vovk:2022}, these guarantees are not immediate in our setting. 
Establishing asymptotic conditional validity is even more challenging. The bounded support of the response induces heteroscedasticity and boundary-driven asymmetry in the conditional distribution, so conditional validity requires regularity conditions tailored to the distribution of the fitted scores rather than a direct application of existing results. Proposition~\ref{prop:conditional} establishes asymptotic conditional validity under regularity conditions.

\medskip

We first outline the framework and notation through the following assumptions. 
\begin{itemize}
\item[A1.] For $i=1,\dots,n$, let $W_i\in(0,1)$ denote the original bounded response and let $Y_i=T(W_i)$ denote the corresponding working-scale response, where $T:(0,1)\to\mathcal Y\subseteq\mathbb R$ is a continuous and strictly increasing mapping. Let $\bX_i\in\mathbb R^p$ denote the covariate vector, and define $\bZ_i=(Y_i,\bX_i)$ and the testing point $\bZ_{\text{new}}=(Y_{\text{new}},\bX_{\text{new}})$. The corresponding realizations are denoted by $\bz_i=(y_i,\bx_i)$ and $y_i=T(w_i)$, where $\bz_i$ and $\bx_i$ denote the corresponding realizations.
\item[A2.] We consider a parametric regression model for the working-scale response, $Y_i\mid \bX_i=\bx_i \sim F(\cdot;\bzeta,\bx_i)$, where $F(\cdot;\bzeta,\bx_i)$ is a continuous CDF supported on $\mathcal Y$, and $\bzeta$ denotes the model parameter vector. Let $\widehat{\bzeta}$ be the estimator of $\bzeta$, computed from the observed sample by a fitting rule that is invariant to permutations of its inputs, such as maximum likelihood estimation. Under possible model misspecification, we denote the probability limit of $\widehat{\bzeta}$ by $\bzeta^\star$, which is assumed to exist and to be an interior point of the parameter space.
\item[A3.] Suppose that the true conditional distribution of $y_i \mid \bx_i$ is denoted by $\Psi(y_i \mid \bx_i)$. Under the correct model specification, $\Psi(y_i \mid \bx_i)=F(y_i; \bzeta^*,\bx_i)$, whereas under the misspecified model, $\Psi(y_i \mid \bx_i) \neq F(y_i; \bzeta^*, \bx_i)$.
\end{itemize}

The above assumptions cover both model frameworks considered in this paper. For beta regression, $T$ is the identity map. Hence, $Y_i=W_i$, $\mathcal Y=(0,1)$, and $F(\cdot;\bzeta,\bx_i)$ is the conditional CDF of the bounded response itself, which corresponds to $F(y_i;\bzeta,\bx_i)=F_{\text{Beta}}(y_i;\mu_i,\phi_i)$, where $\mu_i$ and possibly $\phi_i$ are modelled through covariates. For the logit-normal regression model, $T$ is a strictly increasing map ($\logit$) from $(0,1)$ to an unbounded working scale $\mathbb R$. In this case, $Y_i$ is the transformed response and $F(\cdot;\bzeta,\bx_i)$ denotes the conditional CDF of $Y_i\mid \bX_i=\bx_i$ on the transformed scale, which corresponds to the Gaussian CDF $F(y_i;\bzeta,\bx_i)=\Phi\left(\frac{y_i-\mu_i}{\sigma_i}\right)$ associated with the fitted regression model on the $\logit$ scale.
These mild assumptions suffice to establish finite-sample marginal validity for the conformal prediction intervals produced by Algorithm~\ref{alg:splitCP} (split conformal) and Algorithm~\ref{alg:fullCP} (full conformal), holding in the presence of heteroscedasticity.

\begin{proposition}[Marginal validity]\label{prop:marginal} 
Consider the setting described in Assumptions A1-A3. Let $\widetilde{\mathcal C}^{\text{split}}_{1-\alpha}(\bx_{\text{new}})$ and $\widetilde{\mathcal C}^{\text{full}}_{1-\alpha}(\bx_{\text{new}})$ denote the conformal prediction sets constructed on the working scale $\mathcal{Y}$ for a new covariate $\bx_{\text{new}}$, and let $\mathcal C^{\text{split}}_{1-\alpha}(\bx_{\text{new}})$ and $\mathcal C^{\text{full}}_{1-\alpha}(\bx_{\text{new}})$ be the corresponding prediction sets on the original bounded response scale. Suppose that $(\bZ_1, \ldots, \bZ_{n},\bZ_{new})$ are exchangeable.
Then, for conformal prediction intervals obtained with Algorithm~\ref{alg:splitCP} (split conformal) and Algorithm~\ref{alg:fullCP} (full conformal), and for a non-conformity score $s_i$ that is symmetric (permutation invariant):

\begin{align*}
 \Pr\left(W_{new} \in \mathcal{C}^{\text{full}}_{1-\alpha}(\bX_{new})\right) & \geq 1-\alpha,
\\ \Pr\left(W_{new} \in \mathcal{C}^{\text{split}}_{1-\alpha}(\bX_{new})\right) & \geq 1-\alpha,
\end{align*}
Equivalently, these bounds hold on the working scale with $Y_{\text{new}}\in \widetilde{\mathcal C}_{1-\alpha}(\bx_{\text{new}})$.

\end{proposition}

In both frameworks, the exchangeability requirement of Proposition~\ref{prop:marginal} holds under the standard sampling assumption underlying the prediction problem. Specifically, if the observed pairs $(W_i,\bX_i)$, together with $(W_{\text{new}},\bX_{\text{new}})$, are drawn exchangeably from a common joint distribution $P$, as is routinely assumed in both beta and logit-normal regression, then exchangeability is preserved on the working scale. In our application (Section \ref{sec:applications}), this assumption is plausible because the construction and testing sets are obtained by random partitioning of the same dataset. Although exchangeability cannot be fully verified from the observed data, potential covariate shift can be explored by comparing the covariate distributions across samples; conformal methods under covariate shift are discussed by \citet{tibshirani:2019}.

To extend our results to asymptotic conditional validity, consider now the following conditions.

\begin{itemize}
    \item[C1.] The estimated conditional CDF converges uniformly to its probability limit:
    \begin{equation*}
        \sup_{Y_i, \bX_i} \left|F(Y_i; \bzeta^*,\bX_i) 
        - F(Y_i; \widehat{\bzeta}, \bX_i)\right| 
        = o_p(1), \quad \text{for }i = 1,\dots,n, \text{new},
    \end{equation*}
    where ``$\text{new}$'' denotes the index for a new observation.
    
    \item[C2.] Suppose that the fitted regression model $\widehat{\mathcal M}$ converges in probability to a limit model ${\mathcal M}^{(*)}$. Let $V_i := S(Y_i,\bX_i;{\mathcal M}^{(*)})$ denote the oracle non-conformity score corresponding to ${\mathcal M}^{(*)}$. The conditional score distribution $G_\star(v) := \Pr(V_{\mathrm{new}} \le v \mid \mathbf{X}_{\mathrm{new}} = \mathbf{x}_{\mathrm{new}})$ admits a density $g_\star$ that is positive on a neighbourhood $\mathcal{N}$ of the $(1-\alpha)$-quantile $q^{*}_{1-\alpha} = G_\star^{-1}(1-\alpha)$.
    \item[C3.] For $i = 1,\dots,n$, let $G_i(v) := \Pr(V_i \le v)$ denote the distribution function for $V_i$, and set $m_n(v) := \frac{1}{n}\sum_{i=1}^{n} G_i(v)$. There exists a neighbourhood $\mathcal{N}$ of $q_{1-\alpha}^*$ such that
\begin{equation*}
   m_n(v) \;\longrightarrow\; G_\star(v),   \quad\text{for every } v \in \mathcal{N}.
\end{equation*}
\end{itemize}

The following result shows that these conditions are sufficient to guarantee the asymptotic conditional validity of split and full conformal prediction under the use of the quantile residual score $r^Q_i$.

\begin{proposition}[Asymptotic conditional validity]
\label{prop:conditional}
Consider the setting described in Assumptions A1-A3. Let $\widetilde{\mathcal C}^{\text{split}}_{1-\alpha}(\bx_{\text{new}})$ and $\widetilde{\mathcal C}^{\text{full}}_{1-\alpha}(\bx_{\text{new}})$ denote the conformal prediction sets constructed on the working scale $\mathcal{Y}$ for a new covariate $\bx_{\text{new}}$, and let $\mathcal C^{\text{split}}_{1-\alpha}(\bx_{\text{new}})$ and $\mathcal C^{\text{full}}_{1-\alpha}(\bx_{\text{new}})$ be the corresponding prediction sets on the original bounded response scale. Suppose that $(\bZ_1, \ldots, \bZ_{n},\bZ_{new})$ are independent, and that conditions \emph{C1--C3} are satisfied. 
Then, for conformal prediction intervals obtained with Algorithm~\ref{alg:splitCP} (split conformal) and Algorithm~\ref{alg:fullCP} (full conformal), and for the quantile residual based non-conformity score $r^Q_i$ defined in  \eqref{eq:res_quantile}:
\begin{align*}
    \Pr\!\left(W_{new} \in \mathcal{C}^{\emph{full}}_{1-\alpha}(\bX_{new})
    \mid \bX_{new} = \bx_{new}\right) &\geq 1 - \alpha + o_p(1),\\
    \Pr\!\left(W_{new} \in \mathcal{C}^{\emph{split}}_{1-\alpha}(\bX_{new})
    \mid \bX_{new} = \bx_{new}\right) &= 1 - \alpha + o_p(1).
\end{align*}
Equivalently, these bounds hold on the working scale with $Y_{\text{new}}\in \widetilde{\mathcal C}_{1-\alpha}(\bx_{\text{new}})$.
\end{proposition}

Condition C1 requires the estimated conditional CDF to converge uniformly to its probability limit, in the sense that the maximum discrepancy between $F(Y_i; \widehat{\bzeta}, \bX_i)$ and $F(Y_i; \bzeta^*,\bX_i)$ vanishes in probability. For the beta regression model, sufficient conditions for C1 include a compact parameter space $\Delta \subset \mathbb{R}^{2(p+1)}$ and the existence of constants $M > 0$ and $\varepsilon > 0$ such that $\|\mathbf{X}_i\| \leq M$ and $Y_i \in [\varepsilon, 1-\varepsilon]$ for all $i$. For the logit-normal regression model, sufficient conditions are a compact parameter space $\Delta \subset \mathbb{R}^{2(p+1)}$ and $\|\mathbf{X}_i\| \leq M$, since $Y_i\in\mathbb R$ and the Gaussian CDF is uniformly continuous in its parameters.
For beta regression, Condition C1 is more general than these sufficient conditions, as it permits outcomes arbitrarily close to the boundaries of $(0,1)$.
Condition C2 is a mild regularity requirement ensuring that $G_{\star}$ is strictly increasing at the relevant quantile; it holds for beta distributions whenever the $(1-\alpha)$-quantile of the score lies in the interior of the support.
Condition C3 requires that the conditional score distribution be covariate-invariant in a neighbourhood of the target quantile. That is, that the oracle score $V=\left|\Phi^{-1}(F(Y;\bzeta^*,\bx))\right|$ be asymptotically pivotal there (fixed distribution), and that the averaged score distribution $m_n(v)=\frac1n\sum_{i=1}^n G_i(v)$ converge to this common law $G_\star$. Although C3 is in fact a strong homogeneity condition, and the reason is structural: the empirical quantile $\widehat{q}_{1-\alpha}$ is computed from the calibration scores alone and therefore does not depend on $\bx_{new}$, so it admits a single probability limit. A covariate-free limit can coincide with the conditional target $q_{1-\alpha}^* = G_\star^{-1}(1-\alpha)$ at every test point only if that target is itself covariate-free near the quantile; C3 is precisely this requirement.
Under correct specification it holds automatically for both model frameworks: the probability integral transform $F(Y;\bzeta^*; \bx)$ is $\mathrm{Uniform}(0,1)$ at every $\bX=\bx$, so $V\sim\mathrm{HalfNormal}(0,1)$ irrespective of the covariates, $G_\star$ does not depend on $\bx_{new}$. Under misspecification the pivot breaks, the conditional score law varies with $\bx_{new}$, and C3 becomes a genuine and restrictive assumption, which is a well-known limitation that non-localized conformal prediction attains conditional, as opposed to marginal, validity only under score homogeneity \citep{lei:2014}.

\section{Simulation Study}\label{sec:simulation}

In this section, we conduct a simulation study to evaluate the empirical marginal coverage (at the 90\% nominal level) and the average width of prediction intervals constructed using split and full conformal prediction methods across four modelling frameworks: 
\begin{enumerate}
\item logit-normal regression,
\item heteroscedastic logit-normal regression,
\item beta regression with covariates affecting only the mean ($\mu$) parameter,
\item beta regression with covariates affecting both the mean ($\mu$) and precision ($\phi$) parameters.
\end{enumerate}

The experiment is designed as it follows.

\begin{itemize}

\item {Covariate generation:} The covariate vectors $\bx_i = (x_{i1},x_{i2},x_{i3})^{\top}$, $i=1,\dots,n$, are generated from a multivariate normal distribution: $\bx_i \stackrel{i.i.d.}{\sim} N_3 (\textbf{0},\bm{\Sigma})$, where $\bm{\Sigma}$ is a compound symmetric matrix with $\Sigma_{jk} = 1$ if $j=k$, and $\Sigma_{jk} = 0.5$ if $j \neq k$.

\item {Sample sizes and replications:} Sample sizes considered are $n \in \{50, 100, 500, 1000\}$. For each simulation setting, the procedure is repeated across multiple Monte Carlo replications to evaluate empirical coverage. Specifically, 10,000 replications are used for $n=50$ to improve stability at small sample sizes, while 1,000 replications are used for the other values of $n$. In each replication, a new test point is generated to assess whether the constructed prediction interval contains the true response value and the interval width.

\item {Response generation:} We consider four data-generating mechanisms (S1-S4) corresponding to the four models under evaluation:

\begin{enumerate}

\item \textbf{logit-normal regression model (S1):}  
    To ensure comparability with the beta regression model (S3), the standard deviation $\sigma$ in the transformation model is calibrated to approximately match the marginal mean and variance of a beta distribution with covariate-dependent mean $\mu$. This is achieved by minimizing a loss function that balances squared differences in empirical means and variances between simulated logit-normal and beta responses. The resulting calibration yields approximate pairings of $\sigma = 1.50,0.90,0.63,0.45$ corresponding to $\phi = 2,5,10,20$, respectively. For parameters $\sigma \in \{0.90,0.63,0.45\}$, we use regression coefficients $(\theta_0, \boldsymbol{\theta}^{\top})^{\top} = (0.5,0.4,-0.3,0.3)^{\top}$, while for the higher-variance case $\sigma = 1.50$, we adopt the more conservative setting $(0.4,0.25,-0.2,0.2)^{\top}$. Responses are simulated as
    \[
    w_i = \text{expit}(\theta_0 + \boldsymbol{x}_i^{\top} \boldsymbol{\theta} + \epsilon_i), \quad \epsilon_i \overset{\text{i.i.d.}}{\sim} \mathcal{N}(0, \sigma^2).
    \]
    
    \item \textbf{Heteroscedastic logit-normal regression model (S2):} To ensure that the standard deviation $\sigma_i$ remains within the range of 0.45 and 1,5, we model it as a log-linear function of the covariates:
    \[
    \log(\sigma_i) = \eta_0 + \bx_i^{\top}\bmeta,
    \]
    where $(\eta_0, \bmeta^{\top})^{\top} = (-0.2, 0.06, 0.06, 0.06)^{\top}$ is a fixed coefficient vector. Given that $\bx_i \sim \mathcal{N}_3(\mathbf{0}, \bm{\Sigma})$, it follows that $\log(\sigma_i) \sim \mathcal{N}(\eta_0, \bmeta^{\top} \bm{\Sigma} \bmeta)$. Applying the empirical rule for the normal distribution, we approximate
    \[
    \log(\sigma_i) \in [\mu - 3\sigma, \mu + 3\sigma] \approx [-0.65,\ 0.25],
    \]
    which implies $\sigma_i \in [\exp(-0.65),\ \exp(0.25)] \approx [0.52,\ 1.28] \subset [0.45,1.5]$. For the mean ($\mu$), we use regression coefficients $(\theta_0, \btheta^{\top})^{\top} = (0.4, 0.25, -0.2, 0.2)^{\top}$. The responses are then generated as
    \[
    w_i = \text{expit}(\theta_0 + \bx_i^{\top} \btheta + \epsilon_i),
    \]
    where $\epsilon_i \overset{\text{i.i.d.}}{\sim} \mathcal{N}\left(0,\ \exp\{2(\eta_0 + \bx_i^{\top} \bmeta)\} \right)$.
    
    \item \textbf{Beta regression model with covariate-dependent mean (S3):}  
    To ensure that the conditional mean $\mu_i$ remains well within the unit interval and avoids values near the boundaries (0 and 1), the regression coefficients are chosen to control the scale of the linear predictor. For precision parameters $\phi \in \{5,\ 10,\ 20\}$, we set $(\theta_0, \btheta^{\top})^{\top} = (0.5,\ 0.4,\ -0.3,\ 0.3)^{\top}$. For $\phi = 2$, a more conservative choice $(\theta_0, \btheta^{\top})^{\top} = (0.4,\ 0.25,\ -0.2,\ 0.2)^{\top}$ is used to mitigate the higher variability associated with lower precision. The responses are simulated from a beta distribution with parameters $(\mu_i \phi,\ (1 - \mu_i) \phi)$, where
    \[
    \mu_i = \text{expit}(\theta_0 + \boldsymbol{x}_i^{\top} \btheta).
    \]

\item \textbf{Beta regression model with covariate-dependent mean and dispersion (S4):}  
    In this scenario, the precision parameter $\phi_i$ is modelled as a log-linear function of the covariates to ensure it remains within the range of 2 and 20:
    \[
    \log(\phi_i) = \eta_0 + \bx_i^{\top} \bmeta,
    \]
    with fixed coefficients $(\eta_0, \bmeta^{\top})^{\top} = (1.85, 0.15, 0.15, 0.15)^{\top}$. Given that $\bx_i \sim \mathcal{N}_3(\mathbf{0}, \bm{\Sigma})$, it follows that $\log(\phi_i) \sim \mathcal{N}(\eta_0, \bmeta^{\top} \bm{\Sigma} \bmeta)$. Using the empirical rule for the normal distribution, we approximate
    \[
    \log(\phi_i) \in [\mu - 3\sigma,\ \mu + 3\sigma] \approx [0.74,\ 2.96],
    \]
    which implies $\phi_i \in [\exp(0.74),\ \exp(2.96)] \approx [2.1,\ 19.3] \subset [2,20]$.
    
    The mean is modelled as
    \[
    \mu_i = \text{expit}(\theta_0 + \bx_i^{\top} \btheta),
    \]
    with coefficients $(\theta_0, \btheta^{\top})^{\top} = (0.4, 0.25, -0.2, 0.2)^{\top}$. The responses are then drawn from a beta distribution with parameters $(\mu_i \phi_i,\ (1 - \mu_i) \phi_i)$.

\item \textbf{Model frameworks:}
The construction of conformal prediction intervals follows the models and scores introduced in Sections \ref{sec:regmods} and \ref{sec:CPbetatreg}. For the logit-normal model (M1), the absolute raw residual is used as the non-conformity score, while for the heteroscedastic logit-normal model (M2), the absolute Pearson residual is employed. In the beta regression framework, two modelling strategies are evaluated: modelling with covariate-dependent $\mu$ only under the absolute Pearson residual (M3(P)) and the absolute quantile residual (M3(Q)), and modelling with both covariate-dependent $\mu$ and $\phi$ under the absolute Pearson residual (M4(P)) and the absolute quantile residual (M4(Q)).

\item \textbf{Tolerance and parameter choice for the full conformal:}
In this simulation study, we set $\varepsilon=10^{-4}$. To achieve this, for models M1 and M2, prediction intervals are constructed over the real line using a fixed grid with step size $10^{-4}$. For models M3 and M4, the prediction interval is inside $(0,1)$. Therefore, we solve for the interval numerically with a tolerance of $10^{-4}$, matching with the cases of M1 and M2, rather than constructing a initial grid. $\rho_{\text{grid}}=3$ is set as the grid expansion parameter for the full conformal under logit-normal regression model, as is introduced in Section \ref{subsec:full_construction}.

\end{enumerate}
\end{itemize}

As a benchmark, we construct bootstrap-based prediction intervals approach from the work of \cite{espinheira:2014}, which relies on standardized weighted residuals derived from the score function of the beta regression model. In this technique, we consider the beta regression model based on covariate-dependent $\mu$ and $\phi$ and use the percentile quantiles to construct the interval. Again, to improve stability, 10,000 replications are used for $n=50$ and 1,000 replications for other values of $n$.

\subsection{Simulation Results}

The results for scenarios S1 and S3, under $\sigma = 0.63$ and $\phi = 10$, respectively, together with results for S2 and S4, are presented in Table \ref{tab:simulation}. 
It also includes the empirical results under corresponding settings.
The rest of the simulation results are presented in Table \ref{tab:extra_simulation} of Appendix \ref{app:simulation}. Table \ref{tab:simulation} reveals several key patterns regarding coverage accuracy and interval efficiency across different simulation scenarios and modelling frameworks. First, both split and full conformal prediction (CP) methods achieve empirical coverage rates close to the nominal 90\% level in all scenarios, regardless of whether the model is correctly specified or misspecified. Notably, full conformal demonstrates superior calibration in small samples ($n=50$), more closely approximating the nominal level (for example, 90.4\% vs. 92.4\% for M1 under S1). This aligns with theoretical expectations, as the full CP avoids the additional randomness introduced by sample splitting. However, as the sample size increases, the coverage differences between the split and full CP methods decrease. In some cases, the split CP performs comparably, suggesting that both methods are asymptotically well-calibrated under regular conditions.

\begin{table}[ht]
\centering
\small
\resizebox{\textwidth}{!}{%
\begin{tabular}{ll*{8}{c}}
\toprule
\multirow{2}{*}{\textbf{Scenario}} & \multirow{2}{*}{\textbf{Sample Size}} 
& \multicolumn{2}{c}{\textbf{S1}} 
& \multicolumn{2}{c}{\textbf{S2}} 
& \multicolumn{2}{c}{\textbf{S3}} 
& \multicolumn{2}{c}{\textbf{S4}} \\
\cmidrule(lr){3-4} \cmidrule(lr){5-6} \cmidrule(lr){7-8} \cmidrule(lr){9-10}
& & Split & Full & Split & Full & Split & Full & Split & Full \\
\midrule
\multirow{4}{*}{\textbf{M1}} 
& $n=50$   & 0.924/0.5106 & 0.904/0.4596 & 0.924/0.6491 & 0.904/0.5955 & 0.929/0.5590 & 0.909/0.5005 & 0.924/0.6932 & 0.904/0.6330 \\
& $n=100$  & 0.913/0.4599 & 0.909/0.4482 & 0.902/0.5971 & 0.906/0.5851 & 0.904/0.5002 & 0.905/0.4867 & 0.912/0.6306 & 0.911/0.6193 \\
& $n=500$  & 0.912/0.4403 & 0.909/0.4382 & 0.910/0.5756 & 0.910/0.5735 & 0.916/0.4807 & 0.913/0.4785 & 0.914/0.6131 & 0.916/0.6089 \\
& $n=1000$ & 0.921/0.4388 & 0.913/0.4376 & 0.905/0.5743 & 0.908/0.5728 & 0.911/0.4773 & 0.914/0.4768 & 0.917/0.6093 & 0.911/0.6078 \\
\midrule
\multirow{4}{*}{\textbf{M2}} 
& $n=50$   & 0.925/0.6378 & 0.903/0.4877 & 0.924/0.7366 & 0.902/0.6129 & 0.927/0.6771 & 0.904/0.5263 & 0.926/0.7591 & 0.904/0.6398 \\
& $n=100$  & 0.913/0.4936 & 0.917/0.4596 & 0.909/0.6195 & 0.915/0.5887 & 0.911/0.5402 & 0.911/0.4993 & 0.915/0.6460 & 0.914/0.6192 \\
& $n=500$  & 0.909/0.4414 & 0.906/0.4382 & 0.904/0.5698 & 0.905/0.5668 & 0.912/0.4842 & 0.911/0.4786 & 0.917/0.6038 & 0.916/0.5990 \\
& $n=1000$ & 0.912/0.4402 & 0.912/0.4383 & 0.913/0.5688 & 0.913/0.5670 & 0.911/0.4779 & 0.915/0.4754 & 0.925/0.6003 & 0.911/0.5983 \\
\midrule
\multirow{4}{*}{\textbf{M3(P)}} 
& $n=50$   & 0.924/0.5241 & 0.903/0.4617 & 0.923/0.6700 & 0.900/0.5982 & 0.926/0.5563 & 0.907/0.4912 & 0.924/0.7032 & 0.904/0.6327 \\
& $n=100$  & 0.915/0.4611 & 0.914/0.4463 & 0.909/0.5998 & 0.905/0.5789 & 0.901/0.4897 & 0.907/0.4734 & 0.905/0.6302 & 0.904/0.6136 \\
& $n=500$  & 0.919/0.4361 & 0.914/0.4333 & 0.910/0.5659 & 0.903/0.5625 & 0.918/0.4652 & 0.914/0.4626 & 0.907/0.6044 & 0.910/0.6009 \\
& $n=1000$ & 0.913/0.4341 & 0.915/0.4326 & 0.902/0.5639 & 0.901/0.5615 & 0.911/0.4624 & 0.912/0.4615 & 0.915/0.6008 & 0.917/0.5994 \\
\midrule
\multirow{4}{*}{\textbf{M3(Q)}} 
& $n=50$   & 0.924/0.5095 & 0.905/0.4580 & 0.923/0.6475 & 0.902/0.5927 & 0.926/0.5468 & 0.908/0.4920 & 0.925/0.6906 & 0.905/0.6303 \\
& $n=100$  & 0.916/0.4577 & 0.911/0.4461 & 0.908/0.5927 & 0.906/0.5801 & 0.902/0.4912 & 0.910/0.4786 & 0.905/0.6277 & 0.915/0.6161 \\
& $n=500$  & 0.912/0.4374 & 0.907/0.4354 & 0.909/0.5701 & 0.908/0.5684 & 0.914/0.4718 & 0.911/0.4698 & 0.906/0.6097 & 0.911/0.6064 \\
& $n=1000$ & 0.914/0.4358 & 0.910/0.4348 & 0.912/0.5690 & 0.907/0.5677 & 0.913/0.4690 & 0.918/0.4687 & 0.916/0.6067 & 0.912/0.6050 \\
\midrule
\multirow{4}{*}{\textbf{M4(P)}} 
& $n=50$   & 0.925/0.6425 & 0.902/0.4946 & 0.924/0.7492 & 0.901/0.6243 & 0.927/0.6606 & 0.903/0.5189 & 0.924/0.7549 & 0.904/0.6411 \\
& $n=100$  & 0.912/0.4960 & 0.910/0.4590 & 0.910/0.6242 & 0.905/0.5870 & 0.911/0.5259 & 0.906/0.4876 & 0.907/0.6441 & 0.911/0.6136 \\
& $n=500$  & 0.900/0.4368 & 0.902/0.4334 & 0.906/0.5638 & 0.908/0.5603 & 0.910/0.4713 & 0.909/0.4654 & 0.910/0.5924 & 0.906/0.5864 \\
& $n=1000$ & 0.908/0.4349 & 0.913/0.4324 & 0.905/0.5620 & 0.913/0.5591 & 0.910/0.4649 & 0.914/0.4626 & 0.918/0.5887 & 0.912/0.5860 \\
\midrule
\multirow{4}{*}{\textbf{M4(Q)}} 
& $n=50$   & 0.929/0.6249 & 0.903/0.4849 & 0.926/0.7258 & 0.902/0.6119 & 0.929/0.6522 & 0.904/0.5161 & 0.924/0.7412 & 0.903/0.6344 \\
& $n=100$  & 0.912/0.4883 & 0.916/0.4570 & 0.910/0.6145 & 0.916/0.5854 & 0.910/0.5253 & 0.913/0.4907 & 0.913/0.6364 & 0.912/0.6137 \\
& $n=500$  & 0.905/0.4385 & 0.907/0.4357 & 0.906/0.5661 & 0.906/0.5635 & 0.905/0.4771 & 0.906/0.4721 & 0.913/0.5973 & 0.916/0.5934 \\
& $n=1000$ & 0.914/0.4372 & 0.915/0.4355 & 0.916/0.5654 & 0.913/0.5633 & 0.911/0.4716 & 0.918/0.4694 & 0.922/0.5949 & 0.914/0.5930 \\

\midrule
\multirow{4}{*}{\textbf{Bootstrap}} 
& $n=50$   
& \multicolumn{2}{c}{0.895/0.4738} 
& \multicolumn{2}{c}{0.894/0.5979} 
& \multicolumn{2}{c}{0.895/0.5042} 
& \multicolumn{2}{c}{0.894/0.6189}  \\

& $n=100$  
& \multicolumn{2}{c}{0.907/0.4549} 
& \multicolumn{2}{c}{0.905/0.5821} 
& \multicolumn{2}{c}{0.903/0.4858} 
& \multicolumn{2}{c}{0.908/0.6069}  \\

& $n=500$  
& \multicolumn{2}{c}{0.906/0.4365} 
& \multicolumn{2}{c}{0.905/0.5643} 
& \multicolumn{2}{c}{0.908/0.4711} 
& \multicolumn{2}{c}{0.911/0.5912}  \\

& $n=1000$ 
& \multicolumn{2}{c}{0.906/0.4372} 
& \multicolumn{2}{c}{0.909/0.5654} 
& \multicolumn{2}{c}{0.910/0.4683} 
& \multicolumn{2}{c}{0.911/0.5925}  \\

\bottomrule
\end{tabular}
}
\caption{Simulation: empirical coverage and average interval width (Coverage / Width) for each combination of prediction method and simulation scenario (S1--S4). For conformal prediction models M1--M4(Q), Split and Full denote split and full conformal prediction, respectively. Bootstrap results are reported once for each scenario and sample size and therefore span the Split and Full columns. For S1 and S3, the results are under $\phi = 10$ or $\sigma = 0.63$.}
\label{tab:simulation}
\end{table}

The full CP produces substantially narrower intervals across all settings. This efficiency advantage is most pronounced in smaller samples (for example, width 0.4946 vs. 0.6425 for M4(P) under S1, $n=50$) but persists in larger samples, without compromising coverage. This gain in efficiency arises because the full CP uses the entire dataset for both model fitting and residual calculation, avoiding the step of data splitting that reduces effective sample size in the split CP. This advantage is especially valuable in small-sample regimes, where the calibration set in split CP may be too small to reliably estimate quantiles, leading to over-conservative intervals. Even as the sample size increases, the full CP retains a moderate efficiency benefit due to its more stable and concentrated residual distribution.

Models incorporating heteroscedasticity (M2 and M4) generally produce wider prediction intervals than their simpler counterparts M1 and M3 when the additional variance or dispersion structure is unnecessary or misspecified (for example, M2 presents wider intervals than M1 almost throughout under S1). This supports the idea that unnecessary heteroscedastic modelling can inflate interval width. When the heteroscedastic models are correctly specified, this width penalty is still visible in small samples (for example, S4 at $n=50$: M4 is wider than M3). This behaviour is consistent with the parsimony principle: although M2 and M4 are more flexible, they require estimation of additional variance or dispersion components, which can inflate the conformal score quantiles in finite samples. As the sample size increases, this estimation penalty diminishes, and the benefit of modelling heteroscedasticity becomes more apparent (for example, S2 at $n=1000$: M2 is narrower than M1).

A clear reduction in both split and full CP interval widths is observed across most scenarios as the sample size increases from \( n = 50 \) to \( n = 1000 \). This progressive narrowing performance reflects the improvement in parameter estimation precision by larger samples. For instance, under the scenario S3 using full conformal prediction, the interval width for model M3(P) decreases from 0.4912 at \( n = 50 \) to 0.4615 at \( n = 1000 \), while M4(P) exhibits an even more pronounced reduction from 0.5189 to 0.4626. The greatest relative decrease in width tends to occur between \( n = 50 \) and \( n = 100 \), highlighting the sensitivity of conformal methods to sample size in small-sample regimes. Importantly, these gains in interval efficiency are achieved without compromising marginal coverage, which remains close to the nominal level across all cases. This indicates that increased sample size enhances the stability and concentration of residual-based quantiles, therefore tightening prediction intervals while preserving their statistical validity.

The bootstrap method achieves empirical coverage close to the nominal 90\% level in most scenarios and sample sizes. However, it shows mild under-coverage when $n=50$, in contrast to the conformal methods, which remain close to or above the nominal level. This may reflect the reliance of the bootstrap procedure on model-based resampling approximations, which can underestimate predictive uncertainty in small samples. Since the bootstrap method is based on a beta regression model with covariate-dependent mean and precision, it is naturally aligned with S3 and S4. Under these beta-generated scenarios, it is competitive in terms of interval width; for example, when $n=500$ under S4, its width is 0.5912, compared with 0.5934 for M4(Q). In contrast, under S1 and S2, where the data are generated from logit-normal models, the bootstrap method is applied under model misspecification. In these cases, its intervals are generally slightly wider than those from the proposed conformal methods; for instance, when $n=1000$ under S1, its width is 0.4372, compared with 0.4358 for M3(Q).

Table \ref{tab:simulation_mcse} in Appendix \ref{app:simulation} reports the Monte Carlo errors under 1,000 iterations corresponding to the empirical coverage and width in Table \ref{tab:simulation}. The Monte Carlo standard errors for coverage are uniformly small across all frameworks, scenarios, and sample sizes, ranging from approximately 0.007 to 0.010, confirming that the empirical coverage estimates in Table \ref{tab:simulation} are stable and reliable. The standard errors for interval width are similarly small and decrease with increasing $n$, reflecting the improved concentration in larger samples. Notably, the standard errors for width are larger under heteroscedastic models (M2, M4), especially at $n=50$, consistent with the greater variability in interval construction when dispersion is estimated from limited data. For the bootstrap, errors are of similar magnitude to those of the conformal methods, indicating that the simulation estimates are comparable. Other Monte Carlo errors are reported in Table \ref{tab:extrasimulation_mcerror} of the Appendix.

\section{Applications}\label{sec:applications}

In this section, we analyze a dataset containing information on college-aged women (18–25 years old) at Brigham Young University, originally collected by \citet{Slack:1997} and later analyzed by \citet{Johnson:2021}. One observation with an implausible age entry (age = 1) was removed from the original dataset of $184$ individuals, resulting in a final sample size of $n=183$. This dataset includes the body fat percentage measured by ``Siri's equation'' \citep{Siri:1956} defined as $$\text{Body Fat Percentage} = \left( \frac{495}{\text{Body Density}} \right) - 450.$$ While this equation can theoretically yield values at 0 or 1 (after scaling), such extremes are not biologically plausible. In this dataset, body fat percentages range from 0.0747 to 0.3849, well within the open interval $(0,1)$, which supports the use of beta and transformation regression models. In addition to BMI (Body Mass Index: calculated as weight in kilograms divided by the square of height in meters), weight, height, and age, the dataset contains 14 standardized circumference measurement (converted to centimeters) for specific body regions. Given the modest sample size, we select a subset of eight covariates to enhance model stability and interpretability. For each covariate, we report a brief description along with summary statistics in the form $(\min,\, \operatorname{mean},\, \max)$: \texttt{BMI} - $(15.80,\, 21.60,\, 31.14)$, \texttt{Neck} (the minimal circumference of the neck, measured perpendicular to its long axis) - $(26.00,\, 31.48,\, 36.00)$, \texttt{Chest} (circumference measured in a horizontal plane at the level of the sixth rib, taken at the end of a normal expiration) - $(43.00,\, 85.02 ,\, 100.00)$, \texttt{Hips} (the maximum horizontal circumference around the buttocks) - $(82.50,\, 96.93,\, 115.00)$, \texttt{Waist} (the minimal horizontal measurement, taken at the end of a normal expiration) - $(58.00,\, 69.46,\, 96.00)$, \texttt{Forearm} (the maximum circumference measured perpendicular to the long axis of the forearm) - $(20.00,\, 23.49,\, 29.00)$, \texttt{PThigh} (Proximal Thigh: circumference immediately below the gluteal fold) - $(45.00,\, 57.38,\,69.50)$, and \texttt{Wrist} (circumference measured perpendicular to the long axis of the forearm) - $(13.50,\, 15.65,\, 19.00)$. We focus on this subset of covariates for the following reasons: \texttt{BMI}, \texttt{Hips}, and \texttt{Waist} have been chosen as preferable variables for body fat in multivariable linear models \citep{Johnson:2021}. \texttt{Chest} and \texttt{Neck} measurements provide information about upper body size and can enhance model accuracy, particularly when fat is not primarily distributed around the \texttt{Hips} and \texttt{Waist}. Notably, recent studies have shown that \texttt{Neck} circumference is a robust marker of body adiposity, especially in women \citep{Padilha:2022}. \texttt{Forearm} and \texttt{Wrist} measurements reflect peripheral limb size and offer insights into body frame and muscle development. Finally, \texttt{PThigh} contributes additional information about the lower body and helps balance the representation of upper and lower body characteristics.

\begin{figure}[ht]
  \centering
  \includegraphics[width=\textwidth]{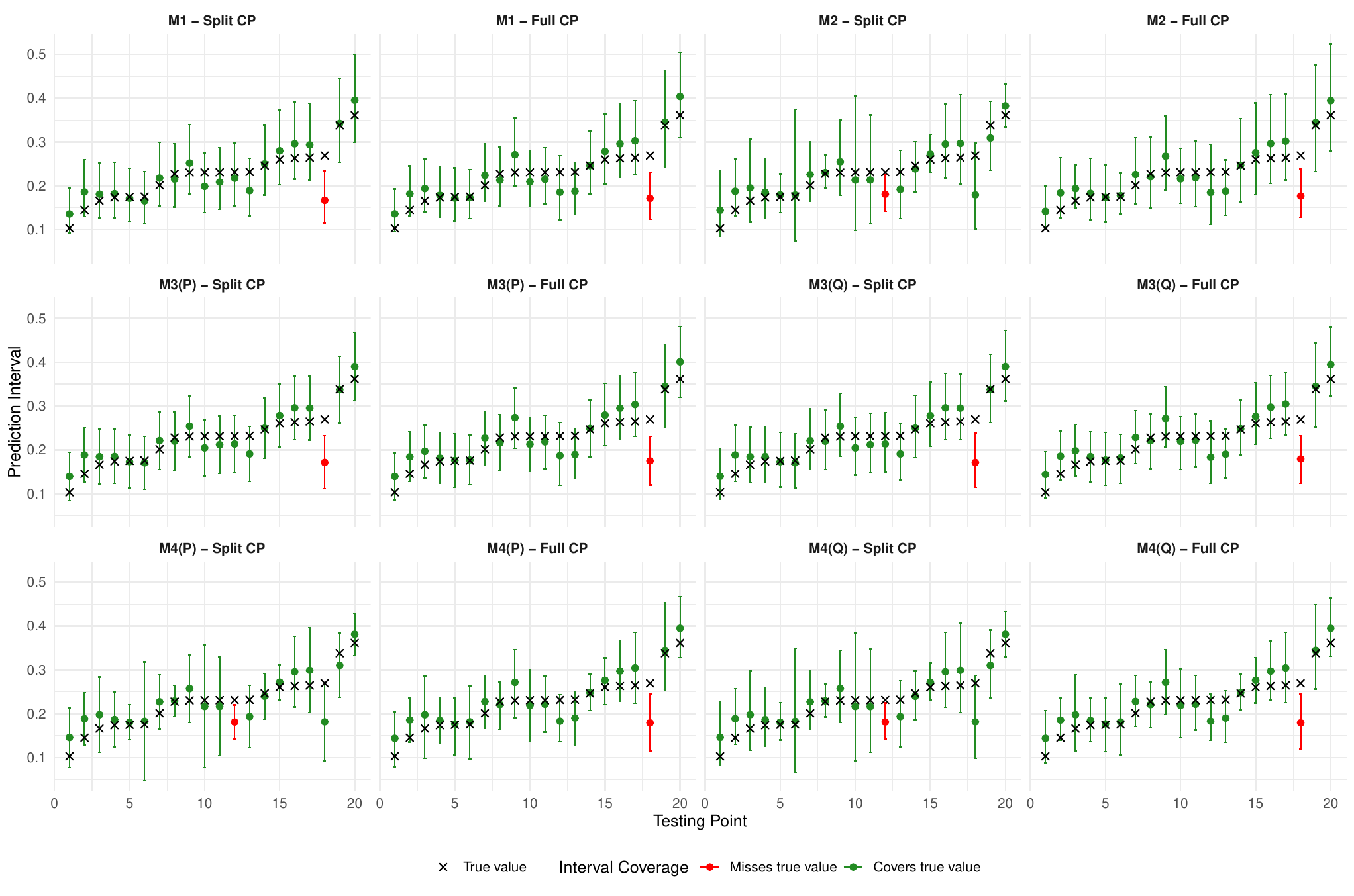}
  \caption{Prediction intervals for all model frameworks.}
  \label{fig:prediction_intervals}
\end{figure}

To implement conformal prediction, the dataset is randomly split into 90\% for interval construction and 10\% reserved as a test set (20 samples) for final evaluation. Within the 90\% construction set, split conformal methods further divide the data equally into training and calibration subsets. Table \ref{tab:application_results} summarizes the empirical coverage and average interval width obtained under the modelling frameworks considered. Figure \ref{fig:prediction_intervals} presents a visualization of prediction intervals among all model frameworks. Each panel corresponds to a different model under either the split or full conformal prediction setting, with test points plotted on the x-axis and predicted intervals shown vertically. Points where the true body fat percentage falls outside the predicted interval are highlighted. Those green or red points are predictions from the transformation regression model or the beta regression model based on the training set (split CP) or the training set plus the calibration set (full CP). Across all settings, conformal prediction intervals consistently achieve the nominal 90\% coverage. As observed in both the simulation study and the real data application, full conformal methods yield slightly narrower intervals than their split counterparts. Introducing heteroscedasticity into either the transformation or beta regression models increases the width of the prediction intervals, as expected. Notably, the beta regression models outperform the transformation models by achieving narrower intervals while maintaining the same level of coverage. Among all configurations, the beta regression model using quantile residuals produces the narrowest intervals.

\begin{table}[ht]
\centering
\small
\setlength{\tabcolsep}{4pt}
\begin{tabular}{ccccc}
\toprule
\multirow{2}{*}{\textbf{Models}} & \multicolumn{2}{c}{\textbf{Split CP}} & \multicolumn{2}{c}{\textbf{Full CP}} \\
\cmidrule(lr){2-3} \cmidrule(lr){4-5}
& Coverage & Width & Coverage & Width \\
\midrule
\textbf{Logit-normal Model - M1} & 0.95 & 0.1457 & 0.95 & 0.1391 \\
\textbf{Heteroscedastic Logit-normal Model - M2} & 0.95 & 0.1599 & 0.95 & 0.1581 \\
\textbf{Beta model ($\mu$, Pearson) - M3(P)} & 0.95 & 0.1323 & 0.95 & 0.1303 \\
\textbf{Beta model ($\mu$, Quantile) - M3(Q)} & 0.95 & 0.1361 & 0.95 & 0.1287 \\
\textbf{Beta model ($\mu$, $\phi$, Pearson) - M4(P)} & 0.95 & 0.1468 & 0.95 & 0.1339 \\
\textbf{Beta model ($\mu$, $\phi$, Quantile) - M4(Q)} & 0.95 & 0.1552 & 0.95 & 0.1287 \\
\textbf{Union interval} & 1 & 0.1887 & 0.95 & 0.1717 \\
\textbf{Intersection interval} & 0.9 & 0.1117 & 0.95 & 0.1137 \\
\midrule
\textbf{Bootstrap} & \multicolumn{4}{c}{Coverage: 0.90; Width: 0.1338} \\
\bottomrule
\end{tabular}
\caption{Body fat data: empirical coverage and average interval width under each prediction model and the bootstrap method (at the 90\% nominal level).}
\label{tab:application_results}
\end{table}

Notably, the missed observations are concentrated around test points 12 or 18 across the different model frameworks. This pattern is robust to both the model specification (transformation \textit{vs.} beta regression) and the CP methodology (split \textit{vs.} full), suggesting these test samples may represent difficult-to-predict individuals rather than isolated model deficiencies. These observations are not located near the extremes of the body fat distribution, indicating that their prediction difficulty is not due to marginal outlyingness in the response. Instead, the missed observations may be conditionally unusual given their covariate profiles. For example, atypical body shape profiles, such as high BMI but low limb circumference, or vice versa, which could result in covariate combinations not well-represented in the dataset used for model construction. This also suggests a connection with multivariable modelling: the non-conformity of these observations may arise from unusual dependence patterns among the body-measurement covariates, rather than from extreme body-fat values alone.

While the bootstrap provides reasonable efficiency under correct model specification, it depends critically on the accuracy of both the mean and dispersion submodels, and relies on asymptotic approximations. 
In contrast, the conformal procedures considered here provide finite-sample marginal coverage guarantees under exchangeability, without requiring correct specification of the working regression model. Notably, the full conformal methods applied to the beta regression framework often yield narrower intervals than the bootstrap, demonstrating both statistical efficiency and robustness. These findings underscore the advantages of full CP, particularly when using quantile-based residuals. 
To assess the consistency of interval estimates across CP models, we perform a sensitivity analysis based on the union and intersection of the prediction intervals. For each test observation, the union interval is obtained by taking the smallest lower endpoint and the largest upper endpoint among all fitted CP intervals. It therefore gives a conservative summary of the range supported by at least one model. The intersection interval is obtained by taking the largest lower endpoint and the smallest upper endpoint among all fitted CP intervals. It represents the region jointly supported by all models, and is empty if the largest lower endpoint exceeds the smallest upper endpoint. The absence of empty intersections confirms a strong level of agreement among the models considered. We report their empirical coverage and average width as a sensitivity analysis of the stability of the prediction intervals across model frameworks and nonconformity scores.

\section{Discussion}\label{sec:discussion}

We study the construction of conformal prediction intervals for continuous bounded outcomes using model-informed non-conformity measures that account for the inherent heteroscedasticity of these data. Because the scores are derived from the fitted model through the probability integral transform, they remain interpretable and, unlike symmetric Pearson-type residuals, they respect the support of the response and account for the boundary-induced asymmetry of the conditional distribution, so that the resulting intervals stay within the admissible range. We establish finite-sample marginal validity for both split and full conformal prediction under exchangeability, together with asymptotic conditional validity under additional regularity conditions. Our simulation study supports these results, with all methods attaining coverage at or near the nominal level, including under model misspecification, and remaining competitive with, or more stable than, a bootstrap benchmark, in particular avoiding the mild under-coverage of the bootstrap in small samples.

A second practical consideration is the trade-off between the two algorithms. Split conformal prediction fits the model once and is essentially instantaneous, making it well suited to large datasets and real-time use, but the data split widens the intervals, especially in small samples. Full conformal prediction uses the entire sample for both fitting and calibration and consistently produces narrower intervals, at the cost of repeated refitting and noticeably higher and more variable runtimes. The adaptive interval-finding routine we employ mitigates this cost, and developing faster approximations to full conformal in this setting is a worthwhile direction. In practice, we recommend the quantile-residual score as the default, given its boundary-respecting behaviour and theoretical interpretability, paired with full conformal when computational resources allow and split conformal when speed or scale is the priority.

We focus on logit-normal regression models and beta regression because of their wide applicability, interpretability, and well-established software implementations. Alternative models and algorithms \citep{mitnik:2013,weinhold:2020,lee:2025} could equally be employed and integrated with the methodology introduced here, and different quantile definitions or normalised conformal prediction methods could be used to further accommodate heteroscedasticity or other features of the data. Other potential extensions include the prediction of multivariate bounded outcomes, conformal methods for zero- and/or one-inflated beta regression models, which pose an additional challenge due to the mixed discrete--continuous nature of the response, and the extension to hierarchical data structures such as spatial, longitudinal, or dynamic settings.

In our real-data application, we also performed a sensitivity analysis comparing the union and intersection of the conformal prediction intervals across methodological frameworks. A natural question follows: since we use four models for bounded responses together with various non-conformity measures, could their predictions be combined, rather than relying on a single model or on the union or intersection of their intervals? One option is to stack the models into an ensemble and aggregate the corresponding prediction sets (see Section~10.4 of \cite{angelopoulos:2024} for a review of stacking methods, and \cite{wu:2026} for a recent proposal), although this requires further analysis to guarantee the validity of the resulting intervals. The real-data application also points to a connection with multivariate modelling. The two non-conforming points we detected did not have marginally extreme responses; rather, they corresponded to atypical \emph{joint} covariate profiles (for example, high BMI together with low limb circumference). This suggests that non-conformity can be driven by the dependence structure of the data rather than by the marginal distribution of the response alone, which is in line with findings that atypical observations are often best identified through explicit modelling of dependence rather than marginal behaviour \citep{deliu:2025}.


\acks{Zhanli Wu was supported by an NMES Faculty Studentship from the Faculty of Natural, Mathematical and Engineering Sciences, King's College London. The authors thank Prof. Vladimir Vovk for helpful discussions about this project that helped them add precision to the results presented here. The authors also thank three reviewers and the Action Editor for their constructive comments provided. }





\newpage

\appendix

\section{Proof of Technical Results}\label{app:proof}

In this appendix, we provide the technical results and proofs supporting the theoretical guarantees in Section~\ref{sec:theory}.

\noindent

\subsection{Lemmas}

The following lemma shows that the empirical quantile convergence is by C1-C3 for continuous regression models, covering beta regression and logit-normal regression models coupled with quantile residuals $r_i^Q$ in \eqref{eq:res_quantile}. 
\begin{lemma}[Quantile convergence]
\label{lem:quantile}
Consider the setting defined by assumptions A1-A3. Define $V_{new} := \left|\Phi^{-1}\!\left(F(Y_{new}; \bzeta^*,\bX_{new})\right)\right|$, where $F$ is a continuous CDF in the first argument.  Define the testing point $G_\star(v) := \Pr(V_{new} \leq v \mid \bX_{new} = \bx_{new})$, and define $q_{1-\alpha}^* := G_\star^{-1}(1-\alpha)$. Denote by $\widehat{q}_{1-\alpha}$ the empirical $(1-\alpha)$-quantile of $\{r_i^Q : i = 1,\ldots,n\}$, and suppose that conditions C1--C3 hold. Then, $\widehat{q}_{1-\alpha} \xrightarrow{\Pr} q_{1-\alpha}^*$ as $n\to\infty$.

\proof

\medskip
Let us denote $U_i := F(Y_i; \bzeta^*,\bX_{new})$ and $\widehat{U}_i := F(Y_i; \widehat{\bzeta},\bX_{new})$, so that $V_i = |\Phi^{-1}(U_i)|$ and $r_i^Q = |\Phi^{-1}(\widehat{U}_i)|$. Let us also denote the oracle and empirical score distribution functions
\begin{equation*}
  \tilde{G}_n(v) := \frac{1}{n}\sum_{i=1}^{n} \mathbf{1}\{V_i \le v\},
  \qquad
  \widehat{G}_n(v) := \frac{1}{n}\sum_{i=1}^{n} \mathbf{1}\{r_i^Q \le v\},
\end{equation*}
and recall $\widehat{q}_{1-\alpha} = \inf\{v : \widehat{G}_n(v) \ge 1-\alpha\}$.

We first prove pointwise convergence of the CDFs $\tilde{G}_n$. Let $v \in \mathcal{N}$. The variables $\mathbf{1}\{V_i \le v\}$ are independent Bernoulli with
\begin{equation*}
    \mathbb{E}\!\left[\mathbf{1}\{V_i \leq v\}\right] = G_i(v), 
    \qquad 
    \mathrm{Var}\!\left(\mathbf{1}\{V_i \leq v\}\right) 
    = G_i(v)\bigl(1 - G_i(v)\bigr) \leq \frac{1}{4}.
\end{equation*}
Then, $\mathbb{E}[\tilde{G}_n(v)] = m_n(v)$ and $\operatorname{Var}(\tilde{G}_n(v)) \le 1/(4n)$.
 By Chebyshev's inequality, for any $t > 0$,
\begin{equation*}
    \Pr\!\left(\left|\tilde{G}_{n}(v) - m_n(v)\right| 
    > t\right) 
    \leq \frac{\mathrm{Var}(\tilde{G}_{n}(v))}{t^2} 
    \leq \frac{1}{4n\,t^2} \longrightarrow 0, \quad n \to \infty.
\end{equation*}
Therefore, $\tilde{G}_{n}(v) - m_n(v) = o_p(1)$. By Condition C3, $m_n(v) - G_\star(v) = o(1)$ for $v \in \mathcal{N}$. Combining the two by the triangle inequality,
\begin{equation}
    \tilde{G}_{n}(v) \xrightarrow{\Pr} G_\star(v) 
    \quad \text{for every } v \in \mathcal{N}.
    \label{eq:pointwise}
\end{equation}

\medskip

Let $\delta_n := \sup_{1\le i\le n}|U_i-\widehat U_i|$, so that $\delta_n = o_p(1)$ by Condition C1.
Fix $v_0\in[q_{1-\alpha}^*-\epsilon, q_{1-\alpha}^*+\epsilon]$ and write $a:=\Phi(-v_0)$, $b:=\Phi(v_0)$, so that $a=1-b$ and $0<a<\frac{1}{2}<b<1$. Let $\delta\in\bigl(0,\,\min\{1-b,\,b-\tfrac12\}\bigr)$ and define $v_0^-:=\Phi^{-1}(b-\delta)$ and $v_0^+:=\Phi^{-1}(b+\delta)$, then $0<v_0^-<v_0<v_0^+$. This implies
\begin{align}
   \{r_i^Q\le v_0\}&=\{a\le \widehat U_i\le b\},\nonumber \\
   \{V_i\le v_0^+\}&=\{a-\delta\le U_i\le b+\delta\},\nonumber \\
   \{V_i\le v_0^-\}&=\{a+\delta\le U_i\le b-\delta\}.
   \label{eq:events}
\end{align}
On the set $E_n(\delta):=\{\delta_n\le\delta\}$ we have $|U_i-\widehat U_i|\le\delta$ for all $i$, then we have the following inclusion of events $\{V_i\le v_0^-\}\subseteq\{r_i^Q\le v_0\}\subseteq\{V_i\le v_0^+\}$. Taking indicator functions and averaging over $i$,
\[
   \tilde G_n(v_0^-)\le \widehat G_n(v_0)\le \tilde G_n(v_0^+)\qquad\text{on }E_n(\delta).
\]

Since $\tilde G_n$ is non-decreasing and $v_0^-\le v_0\le v_0^+$, both $\widehat G_n(v_0)$ and $\tilde G_n(v_0)$ lie in $[\tilde G_n(v_0^-),\tilde G_n(v_0^+)]$ on
$E_n(\delta)$, so there $|\widehat G_n(v_0)-\tilde G_n(v_0)|\le \tilde G_n(v_0^+)-\tilde G_n(v_0^-)$. As $\Phi^{-1}$ is continuous at $b\in(0,1)$ and $G_\star$ at $v_0$
(condition~C2), $G_\star(v_0^+)-G_\star(v_0^-)\to0$ as $\delta\downarrow0$. Fix $\eta>0$. Then, fix $\delta$ (with $v_0^\pm\in\mathcal N$) so small that $G_\star(v_0^+)-G_\star(v_0^-)<\eta$. The points $v_0^\pm$ are now fixed, so \eqref{eq:pointwise} applies at them and $\tilde G_n(v_0^+)-\tilde G_n(v_0^-)\xrightarrow{\Pr}G_\star(v_0^+)-G_\star(v_0^-)<\eta$.
With $\Pr(E_n(\delta))\to1$,
\[
   \Pr\bigl(|\widehat G_n(v_0)-\tilde G_n(v_0)|>\eta\bigr)
   \le \Pr\bigl(E_n(\delta)^c\bigr)
     +\Pr\bigl(\tilde G_n(v_0^+)-\tilde G_n(v_0^-)>\eta\bigr)\longrightarrow0 .
\]
As $\eta>0$ was arbitrary, $\widehat G_n(v_0)-\tilde G_n(v_0)=o_p(1)$.

\medskip

Combining $\widehat G_n(v_0)-\tilde G_n(v_0)=o_p(1)$ with the oracle convergence
\eqref{eq:pointwise} at $v_0$,
\[
   \widehat G_n(v_0)=\tilde G_n(v_0)+o_p(1)\xrightarrow{\Pr}G_\star(v_0).
\]
By Condition C2, $G_\star(q_{1-\alpha}^*-\epsilon)<1-\alpha<G_\star(q_{1-\alpha}^*+\epsilon)$. Since $\widehat G_n$ is non-decreasing and right-continuous, the generalized inverse
$\widehat q_{1-\alpha}=\inf\{v:\widehat G_n(v)\ge1-\alpha\}$ satisfies $\{\widehat q_{1-\alpha}\le v\}=\{\widehat G_n(v)\ge1-\alpha\}$ for every $v$. Then,
\begin{align*}
   \Pr\bigl(\widehat q_{1-\alpha}>q_{1-\alpha}^*+\epsilon\bigr)
   &=\Pr\bigl(\widehat G_n(q_{1-\alpha}^*+\epsilon)<1-\alpha\bigr)\longrightarrow0,\\
   \Pr\bigl(\widehat q_{1-\alpha}\le q_{1-\alpha}^*-\epsilon\bigr)
   &=\Pr\bigl(\widehat G_n(q_{1-\alpha}^*-\epsilon)\ge1-\alpha\bigr)\longrightarrow0,
\end{align*}
the limits hold because $\widehat G_n(q_{1-\alpha}^*+\epsilon)\xrightarrow{\Pr}G_\star(q_{1-\alpha}^*+\epsilon)>1-\alpha$ and
$\widehat G_n(q_{1-\alpha}^*-\epsilon)\xrightarrow{\Pr}G_\star(q_{1-\alpha}^*-\epsilon)<1-\alpha$.
Therefore, $\Pr(|\widehat q_{1-\alpha}-q_{1-\alpha}^*|>\epsilon)\to0$. This holds for every $\epsilon>0$ small enough that $[q_{1-\alpha}^*-\epsilon,\,q_{1-\alpha}^*+\epsilon]\subset\mathcal N$ and $G_\star$ is strictly increasing there; implying the result $\widehat q_{1-\alpha}\xrightarrow{\Pr}q_{1-\alpha}^*$.

\end{lemma}

\subsection{Proof of Proposition \ref{prop:marginal}}

Let $\alpha\in(0,1)$ be the fixed significance level, and define rank indices $k := \lceil (1-\alpha)(n+1)\rceil$ for full CP and $k_{\mathrm{cal}} := \lceil (1-\alpha)(n_{cal}+1)\rceil$ for split CP.

\paragraph{Full CP}
Since $T:(0,1)\to \mathcal{Y}$ is a strictly increasing mapping by 
Assumption A1, it has a well-defined inverse $T^{-1}:\mathcal{Y}\to(0,1)$. By construction, Algorithm \ref{alg:fullCP} searched over $y\in\mathcal{Y}$, so $\widetilde{\mathcal{C}}^{\text{full}}_{1-\alpha}(\bX_{\text{new}})\subseteq\mathcal{Y}$. Since $Y_{\text{new}}=T(W_{\text{new}})$ and $T^{-1}$ is strictly increasing, $\{W_{\text{new}}\in\mathcal{C}^{\text{full}}_{1-\alpha}(\bX_{\text{new}})\}$ if and only if $\{Y_{\text{new}}\in\widetilde{\mathcal C}^{\text{full}}_{1-\alpha}(\bX_{\text{new}})\}$.

Fix the candidate value at the true response, $y=Y_{new}$. 
We show that under assumptions A1-A3, the proof then follows by standard CP theory \citep[Chapter 3]{angelopoulos:2024}. For notational simplicity, in the remainder of the proof we use the index $n+1$ interchangeably with ``new'' to denote the test observation.
The scores $(s_1^{(y)},\dots,s_{new}^{(y)})$ are exchangeable. 
By construction of Algorithm~\ref{alg:fullCP}, the threshold $q_{1-\alpha}^{(y)}$ is the $k$-th smallest of $\{s_1^{(y)},\dots,s_n^{(y)}\}$ (with the convention $q_{1-\alpha}^{(y)}=+\infty$ when $k>n$), and $Y_{new}\in\widetilde{\mathcal{C}}^{\text{full}}_{1-\alpha}(\bX_{new})$ if and only if $s_{new}^{(y)}\le q_{1-\alpha}^{(y)}$. By exchangeability and the quantile lemma \citep[Lemma~1]{tibshirani:2019}, which requires no assumption on ties, it follows that:
\begin{equation*}
  \Pr\!\left(Y_{new}\in\widetilde{\mathcal{C}}^{\text{full}}_{1-\alpha}(\bX_{new})\right) \;\geq\; 1+\alpha.
 \end{equation*}

\paragraph{Convex Hull.} Since the convex hull $\overline{\mathcal C}^{\mathrm{full}}_{1-\alpha}(\bX_{\mathrm{new}})$ of $\mathcal C^{\mathrm{full}}_{1-\alpha}(\bX_{\mathrm{new}})$, reported by Algorithm~\ref{alg:fullCP}, contains it (or is equal to), the same bound holds for $\overline{\mathcal C}^{\mathrm{full}}_{1-\alpha}(\bX_{\mathrm{new}})$.

\medskip

\paragraph{Split CP}
By construction in Algorithm \ref{alg:splitCP}, the model is fitted once, on $\{\bZ_i:i\in\mathcal{I}_{\text{train}}\}$, so the fitted statistic $\widehat{\bzeta}$, and hence the score function
$z\mapsto s(z;\widehat{\mathcal{M}}(\widehat{\bzeta}))$, is a fixed measurable function once $\mathcal{I}_{\text{train}}$ is given. Conditional on $\mathcal{I}_{\text{train}}$, the
calibration responses and the test point, $\{\bZ_i:i\in\mathcal{I}_{\text{cal}}\}\cup\{\bZ_{new}\}$, remain exchangeable and are independent of the training fold; applying the common fixed score function gives exchangeable scores $\{s_i:i\in\mathcal{I}_{\text{cal}}\}\cup\{s_{new}\}$, almost surely tie-free by the assumptions of the Proposition.  Repeating the rank argument aforementioned with $n_{cal}$ in place of $n$, and $k_{cal}$ in place of $k$ gives, conditionally on $\mathcal{I}_{\text{train}}$,
\begin{equation*}
  1-\alpha
  \;\le\;
  \Pr\!\left(Y_{new}\in\widetilde{\mathcal{C}}^{\text{split}}_{1-\alpha}(\bX_{new})\,\middle|\,\mathcal{I}_{\text{train}}\right).
\end{equation*}

Taking expectations over $\mathcal{I}_{\text{train}}$ and applying the law of total expectation yields the unconditional bound $1-\alpha
  \;\le\;
  \Pr\!\left(Y_{new}\in\widetilde{\mathcal{C}}^{\text{split}}_{1-\alpha}(\bX_{new})\right)$, which completes the proof since the equivalence $\{W_{\text{new}}\in\mathcal{C}^{\text{split}}_{1-\alpha}(\bX_{\text{new}})\} \iff \{Y_{\text{new}}\in\widetilde{\mathcal C}^{\text{split}}_{1-\alpha}(\bX_{\text{new}})\}$ holds as before.

\subsection{Proof of Proposition \ref{prop:conditional}}

We prove the result for full CP; the argument for split CP is analogous, with the calibration set $\mathcal{J}_{\text{cal}}$ replacing $\{1,\ldots,n\}$. As is verified in the Proof of Proposition 1, the equivalence $\{W_{\text{new}}\in\mathcal{C}^{\text{full}}_{1-\alpha}(\bX_{\text{new}})\} \iff \{Y_{\text{new}}\in\widetilde{\mathcal C}^{\text{full}}_{1-\alpha}(\bX_{\text{new}})\}$ holds.

Define the conditional coverage gap
\begin{equation*}
    \Delta_n(\bx_{new}) := \Pr\!\left(Y_{new} \in 
    \widetilde{\mathcal{C}}^{\text{full}}_{1-\alpha}(\bX_{new}) 
    \mid \bX_{new} = \bx_{new}\right) - (1-\alpha).
\end{equation*}
We show $\Delta_n(\bx_{new}) = o_p(1)$. 
Define $s_{\text{new}} := S(Y_{\text{new}};\widehat{\bzeta},\bX_{\text{new}})$ as the non-conformity score for the new observation based on the fitted model $\widehat{\mathcal{M}}$, and $V_{\text{new}}:=S(Y_{\text{new}};\bzeta^*,\bX_{\text{new}})$ as the oracle non-conformity score for the new observation on the limit model $\mathcal{M}^{(*)}$. Then, define $R_{new} := s_{new} - V_{new}$. By Condition C1, $R_{new} = o_p(1)$. Denote $G_\star(v) := \Pr(V_{new} \leq v \mid \bX_{new} = \bx_{new})$, and define $q_{1-\alpha}^* := G_\star^{-1}(1-\alpha)$. By construction of Algorithm \ref{alg:fullCP}, $Y_{new} \in \widetilde{\mathcal{C}}^{\text{full}}_{1-\alpha}(\bX_{new})$ if and only if $s_{new} \leq \widehat{q}_{1-\alpha}$, where $\widehat{q}_{1-\alpha}$ is the empirical $(1-\alpha)$-quantile of $\{s_i : i = 1,\ldots,n\}$. Adding and subtracting $\Pr(s_{new} \leq q_{1-\alpha}^* \mid \bx_{new})$,
\begin{align}
    \Delta_n(\bx_{new}) 
    &= \underbrace{\Pr(s_{new} \leq 
    \widehat{q}_{1-\alpha} \mid \bx_{new}) 
    - \Pr(s_{new} \leq q_{1-\alpha}^* \mid 
    \bx_{new})}_{(A)} \nonumber\\
    &+ \underbrace{\Pr(s_{new} \leq q_{1-\alpha}^* \mid 
    \bx_{new}) - (1-\alpha)}_{(B)}.
    \label{eq:decomp}
\end{align}

\medskip
\paragraph{Term (B) vanishes.}
Write $s_{new} = V_{new} + R_{new}$ with $R_{new} = o_p(1)$. For any $\delta > 0$ we bound $\Pr(s_{new} \leq q_{1-\alpha}^* \mid \bx_{new})$ from above and below.

For the upper bound, partitioning on $\{|R_{new}| > \delta\}$ and $\{|R_{new}| \leq \delta\}$ and noting that on $\{|R_{new}| \leq \delta\}$ the inequality $s_{new} \leq q_{1-\alpha}^*$ together 
with $V_{new} = s_{new} - R_{new} \leq s_{new} + \delta$ gives $V_{new} \leq q_{1-\alpha}^* + \delta$,

\begin{equation}
    \{s_{new} \leq q_{1-\alpha}^*\} \cap \{|R_{new}| \leq \delta\} \subseteq \{V_{new} \leq q_{1-\alpha}^* + \delta\}.
    \label{eq:subsets}
\end{equation}

Using $\Pr(B \cap C) \geq \Pr(B) - \Pr(C^c)$ with $B = \{s_{new} \leq q_{1-\alpha}^*\}$ and $C = \{|R_{new}| \leq \delta\}$ together with \eqref{eq:subsets},

\begin{align*}
    \Pr\!\left(s_{new} \leq q_{1-\alpha}^* \mid 
    \bx_{new}\right) 
    &\leq \Pr\!\left(V_{new} \leq q_{1-\alpha}^* + \delta \mid 
    \bx_{new}\right) + \Pr(|R_{new}| > \delta) \\
    &= G_\star(q_{1-\alpha}^* + \delta) + \Pr(|R_{new}| > \delta).
\end{align*}

Similarly, for the lower bound, on $\{|R_{new}| \leq \delta\}$ the inequality $V_{new} \leq q_{1-\alpha}^* - \delta$ implies $s_{new} = V_{new} + R_{new} \leq q_{1-\alpha}^*$, so that
\begin{equation*}
    \{V_{new} \leq q_{1-\alpha}^* - \delta\} 
    \cap \{|R_{new}| \leq \delta\} 
    \subseteq \{s_{new} \leq q_{1-\alpha}^*\}.
\end{equation*}
Now, set $B = \{V_{new} \leq q_{1-\alpha}^* - \delta\}$ and $C = \{|R_{new}| \leq \delta\}$,
\begin{align*}
    \Pr\!\left(s_{new} \leq q_{1-\alpha}^* \mid 
    \bx_{new}\right) 
    &\geq \Pr\!\left(V_{new} \leq q_{1-\alpha}^* - \delta 
    \mid \bx_{new}\right) - \Pr(|R_{new}| > \delta) \\
    &= G_\star(q_{1-\alpha}^* - \delta) - \Pr(|R_{new}| > \delta).
\end{align*}

Combining the two bounds,
\begin{equation*}
    G_\star(q_{1-\alpha}^* - \delta) - \Pr(|R_{new}| > \delta) 
    \leq \Pr\!\left(s_{new} \leq q_{1-\alpha}^* \mid 
    \bx_{new}\right) 
    \leq G_\star(q_{1-\alpha}^* + \delta) + \Pr(|R_{new}| > \delta).
\end{equation*}
Fix $\eta > 0$. By continuity of $G_\star$ at $q_{1-\alpha}^*$ (Condition C2), choose $\delta > 0$ such that $G_\star(q_{1-\alpha}^* + \delta) - G_\star(q_{1-\alpha}^* - 
\delta) \leq \eta/2$, and then $N$ such that $\Pr(|R_{new}| > \delta) \leq \eta/2$ for all $n > N$, which is possible since $R_{new} = o_p(1)$. For $n > N$, both bounds lie within $\eta$ of $G_\star(q_{1-\alpha}^*) = 1 - \alpha$, so that
\begin{equation*}
    \left|\Pr\!\left(s_{new} \leq q_{1-\alpha}^* \mid 
    \bx_{new}\right) - (1-\alpha)\right| \leq \eta.
\end{equation*}
As $\eta > 0$ was arbitrary, $\Pr(s_{new} \leq q_{1-\alpha}^* \mid \bx_{new}) = 1 - \alpha + o(1)$, so that $(B) = o(1)$.

\medskip
\paragraph{Term (A) vanishes.}
By the indicator identity $|\mathbf{1}\{t \leq a\} - \mathbf{1}\{t \leq b\}| = \mathbf{1}\{\min(a,b) < t \leq \max(a,b)\}$,
\begin{equation*}
    |(A)| \leq \Pr\!\left(\min(\widehat{q}_{1-\alpha}, 
    q_{1-\alpha}^*) < s_{new} \leq 
    \max(\widehat{q}_{1-\alpha}, q_{1-\alpha}^*) \mid 
    \bx_{new}\right).
\end{equation*}
Fix $\epsilon > 0$. On $\{|\widehat{q}_{1-\alpha} - q_{1-\alpha}^*| \leq \epsilon\}$ this event is contained in $\{q_{1-\alpha}^* - \epsilon < s_{new} \leq 
q_{1-\alpha}^* + \epsilon\}$, so
\begin{equation*}
    |(A)| \leq \Pr\!\left(q_{1-\alpha}^* - \epsilon < 
    s_{new} \leq q_{1-\alpha}^* + \epsilon \mid 
    \bx_{new}\right) 
    + \Pr\!\left(|\widehat{q}_{1-\alpha} - q_{1-\alpha}^*| 
    > \epsilon\right).
\end{equation*}
Fix $\eta > 0$. By continuity of $G_\star$, choose $\epsilon, \delta > 0$ with $G_\star(q_{1-\alpha}^* + \epsilon + \delta) - G_\star(q_{1-\alpha}^* - \epsilon - \delta) \leq \eta/3$; then, 
arguing via $\{|R_{new}| \leq \delta\}$ as in Term (B),
\begin{equation*}
    \Pr\!\left(q_{1-\alpha}^* - \epsilon < s_{new} 
    \leq q_{1-\alpha}^* + \epsilon \mid \bx_{new}\right) 
    \leq \frac{\eta}{3} + \Pr(|R_{new}| > \delta).
\end{equation*}
Since $R_{new} = o_p(1)$ and $\widehat{q}_{1-\alpha} \xrightarrow{\Pr} q_{1-\alpha}^*$ (Lemma \ref{lem:quantile}), there is $N$ with $\Pr(|R_{new}| > \delta) \leq \eta/3$ and $\Pr(|\widehat{q}_{1-\alpha} - q_{1-\alpha}^*| > \epsilon) \leq \eta/3$ for $n > N$. Hence $|(A)| \leq \eta$ for $n > N$, and as $\eta$ was arbitrary, $|(A)| = o(1)$.

\medskip
Combining the bounds on $(A)$ and $(B)$ gives $\Delta_n(\bx_{new}) = o_p(1)$, which establishes asymptotic conditional validity for full CP. 

\medskip
\paragraph{Convex Hull.} For the convex hull
$\overline{\mathcal C}^{\mathrm{full}}_{1-\alpha}(\bX_{\mathrm{new}})$
reported by Algorithm~\ref{alg:fullCP},
\[
\Pr\!\left(W_{\mathrm{new}}\in
\overline{\mathcal C}^{\mathrm{full}}_{1-\alpha}(\bX_{\mathrm{new}})
\mid \bX_{\mathrm{new}}=\bx_{\mathrm{new}}\right)\ge 1-\alpha+o_p(1),
\]
with equality whenever
$\mathcal C^{\mathrm{full}}_{1-\alpha}(\bX_{\mathrm{new}})$ is connected. The
split conformal set is always an interval, so its statement is unchanged.

\section{Additional Simulation Results}
\label{app:simulation}

During the simulation process for \( n = 50 \) with Monte Carlo iteration \( N = 10{,}000 \) under the split conformal prediction (CP) setting, we encountered occasional failures in model fitting due to numerical instability, particularly in heteroscedastic models. When such cases occurred, the corresponding random seed was discarded and replaced with a new one to ensure the total number of valid replications remained at \( N = 10,000 \). The affected random seeds, which caused model-fitting failures and were excluded from the simulation, are listed below:

\begin{itemize}
  \item \textbf{S1-M2}: $\sigma = 1.5$: 8608, $\sigma = 0.9$: 8608, $\sigma = 0.63$: 8608, $\sigma = 0.45$: 8608.
  \item \textbf{S3-M2}: $\phi = 2$: 3403, $\phi = 5$: 2807.
  \item \textbf{S2-M2}: 3311.
\end{itemize}

For the bootstrap-based prediction intervals, occasional numerical failures were also observed when \(n=50\). The skipped bootstrap seeds were as follows: S1-Bootstrap: 6952; S2-Bootstrap: 9437; S3-Bootstrap: 3052, 6087; and S4-Bootstrap: 3635, 8247, 9180, 9644. 

These issues were limited to a very small number of seeds and did not materially affect the overall validity or stability of the simulation results.

\begin{table}[ht]
\centering
\small
\resizebox{\textwidth}{!}{%
\begin{tabular}{ll*{8}{c}}
\toprule
\multirow{2}{*}{\textbf{Scenario}} & \multirow{2}{*}{\textbf{Sample Size}} 
& \multicolumn{2}{c}{\textbf{S1}} 
& \multicolumn{2}{c}{\textbf{S2}} 
& \multicolumn{2}{c}{\textbf{S3}} 
& \multicolumn{2}{c}{\textbf{S4}} \\
\cmidrule(lr){3-4} \cmidrule(lr){5-6} \cmidrule(lr){7-8} \cmidrule(lr){9-10}
& & Split & Full & Split & Full & Split & Full & Split & Full \\
\midrule
\multirow{4}{*}{\textbf{M1}} 
& $n=50$   & 0.0083/0.0031 & 0.0095/0.0022 & 0.0085/0.0031 & 0.0094/0.0022 & 0.0079/0.0036 & 0.0085/0.0026 & 0.0079/0.0034 & 0.0087/0.0024 \\
& $n=100$  & 0.0089/0.0022 & 0.0091/0.0018 & 0.0094/0.0022 & 0.0092/0.0016 & 0.0093/0.0026 & 0.0093/0.0022 & 0.0090/0.0025 & 0.0090/0.0019\\
& $n=500$  & 0.0090/0.0016 & 0.0091/0.0015 & 0.0091/0.0012 & 0.0091/0.0010 & 0.0088/0.0020 & 0.0089/0.0019 & 0.0089/0.0015 & 0.0088/0.0013 \\
& $n=1000$ & 0.0085/0.0015 & 0.0089/0.0015 & 0.0093/0.0010 & 0.0091/0.0009 & 0.0090/0.0019 & 0.0089/0.0019 & 0.0087/0.0013 & 0.0090/0.0012 \\
\midrule
\multirow{4}{*}{\textbf{M2}} 
& $n=50$   & 0.0087/0.0072 & 0.0092/0.0039 & 0.0091/0.0064 & 0.0091/0.0041 & 0.0073/0.0070 & 0.0092/0.0040 & 0.0086/0.0067 & 0.0091/0.0046 \\
& $n=100$  & 0.0089/0.0041 & 0.0087/0.0024 & 0.0091/0.0044 & 0.0088/0.0028 & 0.0090/0.0043 & 0.0090/0.0026 & 0.0088/0.0049 & 0.0089/0.0038 \\
& $n=500$  & 0.0091/0.0019 & 0.0092/0.0016 & 0.0093/0.0022 & 0.0093/0.0020 & 0.0090/0.0019 & 0.0090/0.0016 & 0.0087/0.0036 & 0.0088/0.0034 \\
& $n=1000$ & 0.0090/0.0017 & 0.0090/0.0015 & 0.0089/0.0020 & 0.0089/0.0018 & 0.0090/0.0015 & 0.0088/0.0013 & 0.0083/0.0034 & 0.0090/0.0033 \\
\midrule
\multirow{4}{*}{\textbf{M3(P)}} 
& $n=50$   & 0.0084/0.0034 & 0.0094/0.0021 & 0.0083/0.0038 & 0.0094/0.0024 & 0.0082/0.0037 & 0.0088/0.0023 & 0.0079/0.0039 & 0.0087/0.0026 \\
& $n=100$  & 0.0088/0.0020 & 0.0089/0.0014 & 0.0091/0.0024 & 0.0093/0.0015 & 0.0094/0.0022 & 0.0092/0.0016 & 0.0093/0.0026 & 0.0093/0.0018 \\
& $n=500$  & 0.0086/0.0010 & 0.0089/0.0009 & 0.0091/0.0010 & 0.0094/0.0007 & 0.0087/0.0013 & 0.0089/0.0011 & 0.0092/0.0012 & 0.0091/0.0009 \\
& $n=1000$ & 0.0089/0.0009 & 0.0088/0.0008 & 0.0094/0.0007 & 0.0094/0.0006 & 0.0090/0.0012 & 0.0090/0.0011 & 0.0088/0.0010 & 0.0087/0.0008 \\
\midrule
\multirow{4}{*}{\textbf{M3(Q)}} 
& $n=50$   & 0.0079/0.0028 & 0.0093/0.0020 & 0.0083/0.0030 & 0.0096/0.0022 & 0.0079/0.0030 & 0.0086/0.0022 & 0.0075/0.0032 & 0.0088/0.0023 \\
& $n=100$  & 0.0088/0.0019 & 0.0090/0.0014 & 0.0091/0.0020 & 0.0092/0.0014 & 0.0094/0.0020 & 0.0091/0.0016 & 0.0093/0.0023 & 0.0088/0.0017 \\
& $n=500$  & 0.0090/0.0011 & 0.0092/0.0010 & 0.0091/0.0010 & 0.0091/0.0008 & 0.0089/0.0013 & 0.0090/0.0012 & 0.0092/0.0012 & 0.0090/0.0010 \\
& $n=1000$ & 0.0089/0.0010 & 0.0091/0.0009 & 0.0090/0.0008 & 0.0092/0.0007 & 0.0089/0.0012 & 0.0087/0.0011 & 0.0088/0.0010 & 0.0090/0.0009 \\
\midrule
\multirow{4}{*}{\textbf{M4(P)}} 
& $n=50$   & 0.0089/0.0075 & 0.0090/0.0042 & 0.0090/0.0066 & 0.0090/0.0043 & 0.0073/0.0072 & 0.0093/0.0039 & 0.0087/0.0068 & 0.0090/0.0048 \\
& $n=100$  & 0.0090/0.0044 & 0.0091/0.0026 & 0.0091/0.0046 & 0.0093/0.0030 & 0.0090/0.0042 & 0.0092/0.0025 & 0.0092/0.0050 & 0.0090/0.0040 \\
& $n=500$  & 0.0095/0.0018 & 0.0094/0.0015 & 0.0092/0.0022 & 0.0091/0.0020 & 0.0091/0.0017 & 0.0091/0.0014 & 0.0091/0.0036 & 0.0092/0.0034 \\
& $n=1000$ & 0.0091/0.0016 & 0.0089/0.0014 & 0.0093/0.0019 & 0.0089/0.0017 & 0.0091/0.0014 & 0.0089/0.0012 & 0.0087/0.0034 & 0.0090/0.0033 \\
\midrule
\multirow{4}{*}{\textbf{M4(Q)}} 
& $n=50$   & 0.0085/0.0071 & 0.0092/0.0040 & 0.0089/0.0062 & 0.0092/0.0041 & 0.0073/0.0067 & 0.0090/0.0038 & 0.0086/0.0064 & 0.0092/0.0046 \\
& $n=100$  & 0.0090/0.0040 & 0.0088/0.0025 & 0.0091/0.0042 & 0.0088/0.0028 & 0.0091/0.0040 & 0.0089/0.0025 & 0.0089/0.0047 & 0.0090/0.0038  \\
& $n=500$  & 0.0093/0.0018 & 0.0092/0.0016 & 0.0092/0.0021 & 0.0092/0.0019 & 0.0093/0.0018 & 0.0092/0.0014 & 0.0089/0.0035 & 0.0088/0.0033 \\
& $n=1000$ & 0.0089/0.0016 & 0.0088/0.0015 & 0.0088/0.0019 & 0.0089/0.0017 & 0.0090/0.0014 & 0.0087/0.0013 & 0.0085/0.0034 & 0.0089/0.0033 \\

\midrule
\multirow{4}{*}{\textbf{Bootstrap}} 
& $n=50$   
& \multicolumn{2}{c}{0.0099/0.0034} 
& \multicolumn{2}{c}{0.0098/0.0036} 
& \multicolumn{2}{c}{0.0095/0.0034} 
& \multicolumn{2}{c}{0.0094/0.0042}  \\

& $n=100$  
& \multicolumn{2}{c}{0.0092/0.0023} 
& \multicolumn{2}{c}{0.0093/0.0027} 
& \multicolumn{2}{c}{0.0094/0.0023} 
& \multicolumn{2}{c}{0.0091/0.0037}  \\

& $n=500$  
& \multicolumn{2}{c}{0.0092/0.0015} 
& \multicolumn{2}{c}{0.0093/0.0020} 
& \multicolumn{2}{c}{0.0091/0.0014} 
& \multicolumn{2}{c}{0.0090/0.0033}  \\

& $n=1000$ 
& \multicolumn{2}{c}{0.0092/0.0015} 
& \multicolumn{2}{c}{0.0091/0.0018} 
& \multicolumn{2}{c}{0.0091/0.0012} 
& \multicolumn{2}{c}{0.0090/0.0033}  \\
\bottomrule
\end{tabular}
}
\caption{Simulation: Monte Carlo errors based on $N=1{,}000$ Monte Carlo replications for the empirical coverage and average interval width estimates reported in Table~\ref{tab:simulation}. Each cell is reported as Monte Carlo error of the empirical coverage/Monte Carlo error of the average interval width. For S1 and S3, the results are under $\phi = 10$ or $\sigma = 0.63$. Note that the corresponding coverage and width for $n=50$ in Table~\ref{tab:simulation} were computed using $N=10{,}000$ replications.}

\label{tab:simulation_mcse}
\end{table}

\begin{table}[ht]
\centering
\resizebox{\textwidth}{!}{%
\begin{tabular}{ll*{12}{c}}
\toprule
\multirow{2}{*}{\textbf{Scenario}} & \multirow{2}{*}{\textbf{Sample Size}} 
& \multicolumn{4}{c}{$\phi=2$, $\sigma=1.5$} 
& \multicolumn{4}{c}{$\phi=5$, $\sigma=0.9$} 
& \multicolumn{4}{c}{$\phi=20$, $\sigma=0.45$} \\
\cmidrule(lr){3-6} \cmidrule(lr){7-10} \cmidrule(lr){11-14}
& & S1-Split & S1-Full & S3-Split & S3-Full 
  & S1-Split & S1-Full & S3-Split & S3-Full 
  & S1-Split & S1-Full & S3-Split & S3-Full \\
\midrule

\multirow{4}{*}{\textbf{M1}} 
& $n=50$   & 0.924/0.8668 & 0.904/0.8371 & 0.925/0.9132 & 0.903/0.8889 
           & 0.924/0.6612 & 0.904/0.6095 & 0.927/0.7242 & 0.903/0.6645
           & 0.924/0.3844 & 0.904/0.3410 & 0.928/0.4149 & 0.911/0.3655 \\
& $n=100$  & 0.913/0.8370 & 0.909/0.8329 & 0.908/0.8875 &  0.903/0.8863
           & 0.913/0.6097 & 0.909/0.5983 & 0.912/0.6654 & 0.917/0.6508
           & 0.913/0.3413 & 0.909/0.3313 & 0.908/0.3654 & 0.908/0.3544 \\
& $n=500$  & 0.912/0.8290 & 0.909/0.8283 & 0.904/0.8825 &  0.913/0.8828 
           & 0.912/0.5902 & 0.909/0.5881 & 0.913/0.6464 & 0.918/0.6439
           & 0.912/0.3247 & 0.909/0.3229 & 0.912/0.3486 & 0.911/0.3466 \\
& $n=1000$ & 0.921/0.8287 & 0.913/0.8283 & 0.913/0.8842 &  0.915/0.8841
           & 0.921/0.5887 & 0.913/0.5875 & 0.911/0.6440 &  0.914/0.6420
           & 0.921/0.3234 & 0.913/0.3223 & 0.905/0.3462 & 0.906/0.3450 \\
\midrule

\multirow{4}{*}{\textbf{M2}} 
& $n=50$   & 0.925/0.8894 & 0.903/0.8442 & 0.920/0.9197 & 0.903/0.8937
           & 0.925/0.7534 & 0.903/0.6343 & 0.928/0.8016 & 0.903/0.6904
           & 0.925/0.5246 & 0.903/0.3667 & 0.926/0.5549 & 0.906/0.3891 \\
& $n=100$  & 0.913/0.8457 & 0.917/0.8381 & 0.906/0.8922 &  0.904/0.8911
           & 0.913/0.6398 & 0.917/0.6099 & 0.911/0.6944 &  0.923/0.6639
           & 0.913/0.3719 & 0.917/0.3409 & 0.914/0.3977 & 0.915/0.3626 \\
& $n=500$  & 0.909/0.8280 & 0.906/0.8274 & 0.902/0.8862 &  0.907/0.8858
           & 0.909/0.5908 & 0.906/0.5878 & 0.920/0.6505 & 0.912/0.6450
           & 0.909/0.3257 &0.906/0.3230  & 0.903/0.3496 & 0.907/0.3462 \\
& $n=1000$ & 0.912/0.8290 & 0.912/0.8285 & 0.909/0.8874 &  0.910/0.8869
           & 0.912/0.5900 & 0.912/0.5882 & 0.914/0.6473 & 0.915/0.6436
           & 0.912/0.3246 & 0.912/0.3229 & 0.912/0.3453 & 0.912/0.3433 \\
\midrule

\multirow{4}{*}{\textbf{M3(P)}} 
& $n=50$   & 0.922/0.8723 & 0.904/0.8320 & 0.925/0.8761 & 0.903/0.8332 
           & 0.924/0.6761 & 0.904/0.6104 & 0.927/0.7027 & 0.905/0.6366 
           & 0.924/0.3915 & 0.905/0.3418 & 0.926/0.4136 & 0.909/0.3608 \\
& $n=100$  & 0.917/0.8335 & 0.915/0.8212 & 0.904/0.8366 & 0.908/0.8241 
           & 0.909/0.6101 & 0.914/0.5927 & 0.910/0.6345 & 0.905/0.6178 
           & 0.914/0.3416 & 0.912/0.3302 & 0.901/0.3606 & 0.905/0.3471 \\
& $n=500$  & 0.915/0.8128 & 0.911/0.8096 & 0.903/0.8192 & 0.913/0.8163 
           & 0.919/0.5811 & 0.916/0.5774 & 0.907/0.6117 & 0.904/0.6076 
           & 0.913/0.3223 & 0.915/0.3205 & 0.915/0.3408 & 0.910/0.3389 \\
& $n=1000$ & 0.910/0.8111 & 0.914/0.8090 & 0.907/0.8194 & 0.906/0.8183 
           & 0.909/0.5789 & 0.914/0.5769 & 0.900/0.6069 & 0.908/0.6054 
           & 0.912/0.3210 & 0.912/0.3199 & 0.915/0.3392 & 0.914/0.3378 \\
\midrule

\multirow{4}{*}{\textbf{M3(Q)}} 
& $n=50$   & 0.924/0.8662 & 0.904/0.8336 & 0.927/0.9041 & 0.904/0.8749 
           & 0.925/0.6593 & 0.904/0.6059 & 0.926/0.7065 & 0.904/0.6492 
           & 0.925/0.3838 & 0.905/0.3404 & 0.926/0.4067 & 0.910/0.3607 \\
& $n=100$  & 0.910/0.8338 & 0.916/0.8281 & 0.904/0.8766 & 0.902/0.8728 
           & 0.915/0.6057 & 0.914/0.5938 & 0.915/0.6506 & 0.914/0.6374 
           & 0.913/0.3400 & 0.911/0.3304 & 0.910/0.3591 & 0.904/0.3489 \\
& $n=500$  & 0.912/0.8238 & 0.906/0.8230 & 0.910/0.8706 & 0.913/0.8695 
           & 0.915/0.5850 & 0.909/0.5830 & 0.917/0.6324 & 0.918/0.6297 
           & 0.912/0.3231 & 0.906/0.3214 & 0.911/0.3432 & 0.915/0.3415 \\
& $n=1000$ & 0.917/0.8237 & 0.910/0.8231 & 0.912/0.8713 & 0.912/0.8710 
           & 0.915/0.5835 & 0.910/0.5825 & 0.916/0.6298 & 0.920/0.6279 
           & 0.914/0.3216 & 0.910/0.3209 & 0.911/0.3411 & 0.911/0.3401 \\
\midrule

\multirow{4}{*}{\textbf{M4(P)}} 
& $n=50$   & 0.923/0.8872 & 0.902/0.8389 & 0.927/0.8891 & 0.907/0.8363 
           & 0.924/0.7519 & 0.902/0.6360 & 0.926/0.7684 & 0.902/0.6553 
           & 0.927/0.5311 & 0.903/0.3732 & 0.925/0.5493 & 0.906/0.3898 \\
& $n=100$  & 0.908/0.8427 & 0.917/0.8253 & 0.904/0.8411 & 0.906/0.8236 
           & 0.907/0.6402 & 0.912/0.6038 & 0.910/0.6540 & 0.906/0.6272 
           & 0.912/0.3739 & 0.917/0.3411 & 0.905/0.3912 & 0.914/0.3593 \\
& $n=500$  & 0.905/0.8117 & 0.907/0.8083 & 0.903/0.8200 & 0.914/0.8162 
           & 0.906/0.5810 & 0.908/0.5769 & 0.904/0.6143 & 0.908/0.6080 
           & 0.902/0.3237 & 0.906/0.3210 & 0.904/0.3448 & 0.908/0.3408 \\
& $n=1000$ & 0.906/0.8107 & 0.912/0.8081 & 0.907/0.8198 & 0.902/0.8178 
           & 0.907/0.5788 & 0.914/0.5757 & 0.907/0.6096 & 0.906/0.6064 
           & 0.910/0.3223 & 0.912/0.3203 & 0.920/0.3401 & 0.916/0.3383 \\
\midrule

\multirow{4}{*}{\textbf{M4(Q)}} 
& $n=50$   & 0.925/0.8830 & 0.902/0.8406 & 0.924/0.9099 & 0.904/0.8779 
           & 0.927/0.7364 & 0.903/0.6284 & 0.927/0.7721 & 0.901/0.6694 
           & 0.930/0.5197 & 0.903/0.3661 & 0.927/0.5410 & 0.906/0.3846 \\
& $n=100$  & 0.910/0.8419 & 0.913/0.8327 & 0.904/0.8780 & 0.905/0.8742 
           & 0.912/0.6313 & 0.914/0.6037 & 0.912/0.6694 & 0.917/0.6450 
           & 0.910/0.3693 & 0.915/0.3401 & 0.906/0.3893 & 0.910/0.3590 \\
& $n=500$  & 0.906/0.8225 & 0.904/0.8217 & 0.901/0.8711 & 0.908/0.8695 
           & 0.906/0.5852 & 0.905/0.5822 & 0.908/0.6348 & 0.916/0.6299 
           & 0.905/0.3245 & 0.909/0.3221 & 0.903/0.3462 & 0.914/0.3431 \\
& $n=1000$ & 0.914/0.8234 & 0.915/0.8227 & 0.912/0.8719 & 0.911/0.8711 
           & 0.913/0.5841 & 0.915/0.5823 & 0.916/0.6321 & 0.919/0.6289 
           & 0.913/0.3232 & 0.915/0.3217 & 0.918/0.3426 & 0.917/0.3406 \\
\bottomrule
\end{tabular}
}
\caption{Coverage and average interval width (Coverage / Width) for each model–scenario combination across $(\phi, \sigma)$ settings, sample sizes, and conformal prediction methods.}
\label{tab:extra_simulation}
\end{table}

\begin{table}[ht]
\centering
\resizebox{\textwidth}{!}{%
\begin{tabular}{ll*{12}{c}}
\toprule
\multirow{2}{*}{\textbf{Scenario}} & \multirow{2}{*}{\textbf{Sample Size}} 
& \multicolumn{4}{c}{$\phi=2$, $\sigma=1.5$} 
& \multicolumn{4}{c}{$\phi=5$, $\sigma=0.9$} 
& \multicolumn{4}{c}{$\phi=20$, $\sigma=0.45$} \\
\cmidrule(lr){3-6} \cmidrule(lr){7-10} \cmidrule(lr){11-14}
& & S1-Split & S1-Full & S3-Split & S3-Full 
  & S1-Split & S1-Full & S3-Split & S3-Full 
  & S1-Split & S1-Full & S3-Split & S3-Full \\
\midrule

\multirow{4}{*}{\textbf{M1}} 
& $n=50$   & 0.0083/0.0025 & 0.0095/0.0018 & 0.0081/0.0028 & 0.0090/0.0023 & 0.0083/0.0034 & 0.0095/0.0025 & 0.0082/0.0043 & 0.0085/0.0032 & 0.0083/0.0026 & 0.0095/0.0018 & 0.0081/0.0029 & 0.0087/0.0020 \\
& $n=100$  & 0.0089/0.0019 & 0.0091/0.0013 & 0.0091/0.0023 & 0.0094/0.0017 & 0.0089/0.0026 & 0.0091/0.0020 & 0.0090/0.0033 & 0.0087/0.0027 & 0.0089/0.0018 & 0.0091/0.0014 & 0.0091/0.0019 & 0.0091/0.0016 \\
& $n=500$  & 0.0090/0.0009 & 0.0091/0.0008 & 0.0093/0.0015 & 0.0089/0.0013 & 0.0090/0.0018 & 0.0091/0.0017 & 0.0089/0.0026 & 0.0087/0.0024 & 0.0090/0.0013 & 0.0091/0.0012 & 0.0090/0.0015 & 0.0090/0.0014 \\
& $n=1000$ & 0.0085/0.0008 & 0.0089/0.0007 & 0.0089/0.0013 & 0.0088/0.0012 & 0.0085/0.0017 & 0.0089/0.0016 & 0.0090/0.0025 & 0.0089/0.0024 & 0.0085/0.0012 & 0.0089/0.0012 & 0.0093/0.0014 & 0.0092/0.0014 \\
\midrule

\multirow{4}{*}{\textbf{M2}} 
& $n=50$   & 0.0087/0.0046 & 0.0092/0.0029 & 0.0083/0.0045 & 0.0082/0.0028  & 0.0087/0.0064 & 0.0092/0.0040 & 0.0078/0.0060 & 0.0091/0.0041 & 0.0087/0.0075 & 0.0092/0.0034 & 0.0077/0.0076 & 0.0089/0.0035 \\
& $n=100$  & 0.0089/0.0034 & 0.0087/0.0020 & 0.0092/0.0031 & 0.0093/0.0019 & 0.0089/0.0044 & 0.0087/0.0027 & 0.0090/0.0045 & 0.0084/0.0030 & 0.0089/0.0035 & 0.0087/0.0020 & 0.0089/0.0036 & 0.0088/0.0020 \\
& $n=500$  & 0.0091/0.0014 & 0.0092/0.0011 & 0.0094/0.0015 & 0.0092/0.0011 & 0.0091/0.0021 & 0.0092/0.0018 & 0.0086/0.0023 & 0.0090/0.0018 & 0.0091/0.0015 & 0.0092/0.0013 & 0.0094/0.0014 & 0.0092/0.0012 \\
& $n=1000$ & 0.0090/0.0011 & 0.0090/0.0008 & 0.0091/0.0011 & 0.0091/0.0009 & 0.0090/0.0019 & 0.0090/0.0017 & 0.0089/0.0018 & 0.0088/0.0015 & 0.0090/0.0014 & 0.0090/0.0012 & 0.0090/0.0012 & 0.0090/0.0010 \\
\midrule

\multirow{4}{*}{\textbf{M3(P)}} 
& $n=50$   & 0.0082/0.0034 & 0.0089/0.0027 & 0.0080/0.0040 & 0.0094/0.0036 & 0.0080/0.0041 & 0.0091/0.0027 & 0.0083/0.0047 & 0.0089/0.0033 & 0.0089/0.0015 & 0.0093/0.0016 & 0.0081/0.0027 & 0.0087/0.0016 \\
& $n=100$  & 0.0087/0.0027 & 0.0088/0.0020 & 0.0093/0.0034 & 0.0091/0.0029 & 0.0091/0.0026 & 0.0089/0.0019 & 0.0091/0.0032 & 0.0093/0.0025 & 0.0089/0.0015 & 0.0090/0.0011 & 0.0094/0.0016 & 0.0093/0.0011 \\
& $n=500$  & 0.0088/0.0013 & 0.0090/0.0011 & 0.0094/0.0025 & 0.0089/0.0023 & 0.0086/0.0014 & 0.0088/0.0012 & 0.0092/0.0021 & 0.0093/0.0019 & 0.0089/0.0008 & 0.0088/0.0007 & 0.0088/0.0009 & 0.0091/0.0008 \\
& $n=1000$ & 0.0091/0.0011 & 0.0089/0.0010 & 0.0092/0.0023 & 0.0092/0.0022 & 0.0091/0.0012 & 0.0089/0.0011 & 0.0095/0.0019 & 0.0091/0.0019 & 0.0090/0.0007 & 0.0090/0.0007 & 0.0088/0.0008 & 0.0089/0.0008 \\
\midrule

\multirow{4}{*}{\textbf{M3(Q)}} 
& $n=50$   & 0.0080/0.0023 & 0.0096/0.0019 & 0.0077/0.0025 & 0.0090/0.0023 & 0.0078/0.0031 & 0.0094/0.0023 & 0.0080/0.0035 & 0.0087/0.0027 & 0.0079/0.0023 & 0.0093/0.0016 & 0.0080/0.0024 & 0.0085/0.0016 \\
& $n=100$  & 0.0091/0.0018 & 0.0088/0.0013 & 0.0093/0.0020 & 0.0094/0.0017 & 0.0088/0.0022 & 0.0089/0.0017 & 0.0088/0.0026 & 0.0089/0.0021 & 0.0089/0.0015 & 0.0090/0.0011 & 0.0091/0.0015 & 0.0093/0.0012 \\
& $n=500$  & 0.0090/0/0009 & 0.0092/0.0007 & 0.0091/0.0013 & 0.0089/0.0012 & 0.0088/0.0013 & 0.0091/0.0012 & 0.0087/0.0018 & 0.0087/0.0016 & 0.0090/0.0008 & 0.0092/0.0007 & 0.0090/0.0010 & 0.0088/0.0008 \\
& $n=1000$ & 0.0087/0.0007 & 0.0091/0.0006 & 0.0090/0.0011 & 0.0090/0.0011 & 0.0088/0.0012 & 0.0091/0.0011 & 0.0088/0.0016 & 0.0086/0.0016 & 0.0089/0.0008 & 0.0091/0.0007 & 0.0090/0.0008 & 0.0090/0.0008 \\
\midrule

\multirow{4}{*}{\textbf{M4(P)}} 
& $n=50$   & 0.0088/0.0046 & 0.0088/0.0036 & 0.0079/0.0049 & 0.0089/0.0041 & 0.0089/0.0067 & 0.0089/0.0044 & 0.0083/0.0063 & 0.0096/0.0044 & 0.0088/0.0079 & 0.0091/0.0037 & 0.0079/0.0077 & 0.0089/0.0035 \\
& $n=100$  & 0.0091/0.0037 & 0.0087/0.0027 & 0.0093/0.0040 & 0.0092/0.0032 & 0.0092/0.0047 & 0.0090/0.0030 & 0.0091/0.0046 & 0.0092/0.0032 & 0.0090/0.0037 & 0.0087/0.0021 & 0.0093/0.0035 & 0.0089/0.0020 \\
& $n=500$  & 0.0093/0.0018 & 0.0092/0.0015 & 0.0094/0.0026 & 0.0089/0.0024 & 0.0092/0.0021 & 0.0091/0.0018 & 0.0093/0.0024 & 0.0091/0.0021 & 0.0094/0.0014 & 0.0092/0.0012 & 0.0093/0.0013 & 0.0091/0.0010 \\
& $n=1000$ & 0.0092/0.0015 & 0.0090/0.0013 & 0.0092/0.0023 & 0.0094/0.0023 & 0.0092/0.0019 & 0.0089/0.0016 & 0.0092/0.0021 & 0.0092/0.0020 & 0.0091/0.0013 & 0.0090/0.0012 & 0.0086/0.0010 & 0.0088/0.0008 \\
\midrule

\multirow{4}{*}{\textbf{M4(Q)}} 
& $n=50$   & 0.0087/0.0041 & 0.0092/0.0030 & 0.0079/0.0038 & 0.0088/0.0030 & 0.0085/0.0062 & 0.0093/0.0041 & 0.0081/0.0058 & 0.0093/0.0041 & 0.0084/0.0076 & 0.0093/0.0035 & 0.0075/0.0074 & 0.0090/0.0034 \\
& $n=100$  & 0.0091/0.0031 & 0.0089/0.0021 & 0.0093/0.0030 & 0.0093/0.0022 & 0.0090/0.0042 & 0.0089/0.0028 & 0.0090/0.0042 & 0.0087/0.0029 & 0.0091/0.0034 & 0.0088/0.0020 & 0.0092/0.0034 & 0.0091/0.0019 \\
& $n=500$  & 0.0092/0.0014 & 0.0093/0.0011 & 0.0094/0.0016 & 0.0091/0.0014 & 0.0092/0.0021 & 0.0093/0.0018 & 0.0091/0.0021 & 0.0088/0.0019 & 0.0093/0.0014 & 0.0091/0.0013 & 0.0094/0.0013 & 0.0089/0.0010 \\
& $n=1000$ & 0.0089/0.0011 & 0.0088/0.0009 & 0.0090/0.0013 & 0.0090/0.0012 & 0.0089/0.0019 & 0.0088/0.0017 & 0.0088/0.0018 & 0.0086/0.0017 & 0.0089/0.0013 & 0.0088/0.0012 & 0.0087/0.0010 & 0.0087/0.0009 \\
\bottomrule
\end{tabular}
}
\caption{Monte Carlo errors based on $N=1{,}000$ Monte Carlo replications for the empirical coverage and average interval width estimates reported in Table~\ref{tab:extra_simulation}. Each cell is reported as Monte Carlo error of the empirical coverage/Monte Carlo error of the average interval width. Note that the corresponding coverage and width estimates for $n=50$ in Table~\ref{tab:extra_simulation} were computed using $N=10{,}000$ replications.}

\label{tab:extrasimulation_mcerror}
\end{table}

All non-bootstrap simulations were conducted in R version 4.3.1 (2023-06-16) on a machine running macOS (Darwin), equipped with an Apple M2 Pro processor and 16 GB of RAM, using a single CPU core. The bootstrap-based simulations, which required substantially greater computational resources, were carried out on the CREATE high-performance computing cluster at King’s College London using Slurm batch jobs in R version 4.5.1. For these bootstrap jobs, we requested 1 node, 1 task, 8 CPU cores, and 32 GB of memory from the CPU partition.


\begin{table}[ht]
\centering
\small
\resizebox{\textwidth}{!}{%
\begin{tabular}{ll*{8}{c}}
\toprule
\multirow{2}{*}{\textbf{Scenario}} & \multirow{2}{*}{\textbf{Sample Size}} 
& \multicolumn{2}{c}{\textbf{S1}} 
& \multicolumn{2}{c}{\textbf{S2}} 
& \multicolumn{2}{c}{\textbf{S3}} 
& \multicolumn{2}{c}{\textbf{S4}} \\
\cmidrule(lr){3-4} \cmidrule(lr){5-6} \cmidrule(lr){7-8} \cmidrule(lr){9-10}
& & Split & Full & Split & Full & Split & Full & Split & Full \\
\midrule

\multirow{4}{*}{\textbf{M1}} 
& $n=50$   & 0.001/0.001 & 0.029/0.004 & 0.001/0.001 & 0.033/0.005 & 0.001/0.001 & 0.031/0.006 & 0.001/0.001 & 0.031/0.005 \\
& $n=100$  & 0.001/0.001 & 0.030/0.005 & 0.001/0.001 & 0.032/0.006 & 0.001/0.001 & 0.031/0.005 & 0.001/0.001 & 0.031/0.005 \\
& $n=500$  & 0.001/0.001 & 0.034/0.008 & 0.001/0.001 & 0.035/0.006 & 0.001/0.001 & 0.035/0.006 & 0.001/0.001 & 0.035/0.005 \\
& $n=1000$ & 0.001/0.001 & 0.039/0.008 & 0.001/0.001 & 0.038/0.006 & 0.001/0.001 & 0.040/0.007 & 0.001/0.001 & 0.039/0.006 \\
\midrule

\multirow{4}{*}{\textbf{M2}} 
& $n=50$   & 0.008/0.003 & 0.251/0.047 & 0.009/0.004 & 0.257/0.041 & 0.009/0.007 & 0.258/0.049 & 0.009/0.003 & 0.260/0.045 \\
& $n=100$  & 0.007/0.003 & 0.236/0.025 & 0.007/0.002 & 0.246/0.032 & 0.007/0.003 & 0.249/0.041 & 0.007/0.002 & 0.258/0.039 \\
& $n=500$  & 0.008/0.003 & 0.310/0.034 & 0.008/0.003 & 0.328/0.040 & 0.008/0.003 & 0.329/0.053 & 0.008/0.003 & 0.360/0.057 \\
& $n=1000$ & 0.008/0.003 & 0.411/0.040 & 0.009/0.003 & 0.449/0.046 & 0.009/0.003 & 0.447/0.054 & 0.009/0.003 & 0.458/0.049 \\
\midrule

\multirow{4}{*}{\textbf{M3(P)}} 
& $n=50$   & 0.005/0.002 & 0.107/0.016 & 0.005/0.002 & 0.112/0.022 & 0.005/0.001 & 0.105/0.028 & 0.005/0.002 & 0.112/0.015 \\
& $n=100$  & 0.005/0.002 & 0.120/0.025 & 0.005/0.002 & 0.135/0.028 & 0.005/0.001 & 0.121/0.012 & 0.005/0.002 & 0.140/0.012 \\
& $n=500$  & 0.008/0.004 & 0.278/0.012 & 0.009/0.004 & 0.307/0.045 & 0.008/0.003 & 0.289/0.031 & 0.010/0.004 & 0.378/0.064 \\
& $n=1000$ & 0.012/0.003 & 0.490/0.034 & 0.013/0.003 & 0.615/0.055 & 0.013/0.003 & 0.517/0.061 & 0.016/0.004 & 0.759/0.113 \\
\midrule

\multirow{4}{*}{\textbf{M3(Q)}} 
& $n=50$   & 0.005/0.003 & 0.106/0.018 & 0.005/0.002 & 0.104/0.010 & 0.005/0.002 & 0.102/0.025 & 0.005/0.002 & 0.112/0.027 \\
& $n=100$  & 0.005/0.002 & 0.117/0.017 & 0.005/0.002 & 0.132/0.016 & 0.005/0.002 & 0.120/0.015 & 0.005/0.002 & 0.141/0.016 \\
& $n=500$  & 0.009/0.007 & 0.280/0.022 & 0.009/0.003 & 0.308/0.047 & 0.008/0.003 & 0.286/0.030 & 0.010/0.003 & 0.381/0.064 \\
& $n=1000$ & 0.012/0.003 & 0.495/0.038 & 0.013/0.003 & 0.616/0.057 & 0.012/0.003 & 0.513/0.059 & 0.016/0.004 & 0.759/0.119 \\
\midrule

\multirow{4}{*}{\textbf{M4(P)}} 
& $n=50$   & 0.007/0.003 & 0.146/0.047 & 0.006/0.002 & 0.131/0.026 & 0.007/0.004 & 0.128/0.021 & 0.007/0.003 & 0.134/0.028 \\
& $n=100$  & 0.006/0.002 & 0.151/0.027 & 0.006/0.002 & 0.149/0.010 & 0.006/0.002 & 0.149/0.031 & 0.007/0.002 & 0.148/0.010 \\
& $n=500$  & 0.011/0.003 & 0.368/0.031 & 0.011/0.003 & 0.374/0.023 & 0.011/0.003 & 0.366/0.025 & 0.011/0.003 & 0.389/0.047 \\
& $n=1000$ & 0.018/0.007 & 0.630/0.045 & 0.016/0.004 & 0.685/0.041 & 0.016/0.003 & 0.668/0.040 & 0.017/0.004 & 0.709/0.070 \\
\midrule

\multirow{4}{*}{\textbf{M4(Q)}} 
& $n=50$   & 0.007/0.004 & 0.131/0.026 & 0.006/0.002 & 0.136/0.032 & 0.006/0.002 & 0.132/0.030 & 0.007/0.003 & 0.135/0.017 \\
& $n=100$  & 0.006/0.002 & 0.152/0.026 & 0.006/0.002 & 0.152/0.009 & 0.006/0.002 & 0.152/0.022 & 0.007/0.002 & 0.159/0.025 \\
& $n=500$  & 0.011/0.003 & 0.366/0.026 & 0.011/0.003 & 0.387/0.027 & 0.011/0.003 & 0.376/0.025 & 0.011/0.003 & 0.385/0.032 \\
& $n=1000$ & 0.016/0.004 & 0.652/0.043 & 0.016/0.003 & 0.705/0.053 & 0.016/0.003 & 0.678/0.037 & 0.016/0.003 & 0.704/0.056 \\

\midrule
\multirow{4}{*}{\textbf{Bootstrap}} 
& $n=50$   
& \multicolumn{2}{c}{10.915/0.832} 
& \multicolumn{2}{c}{11.918/0.798} 
& \multicolumn{2}{c}{12.486/0.918} 
& \multicolumn{2}{c}{12.307/1.293}  \\

& $n=100$  
& \multicolumn{2}{c}{11.475/0.697} 
& \multicolumn{2}{c}{12.556/0.374} 
& \multicolumn{2}{c}{13.034/0.508} 
& \multicolumn{2}{c}{15.061/0.903}  \\

& $n=500$  
& \multicolumn{2}{c}{23.058/0.438} 
& \multicolumn{2}{c}{25.045/1.087} 
& \multicolumn{2}{c}{23.101/0.251} 
& \multicolumn{2}{c}{23.553/0.236}  \\

& $n=1000$ 
& \multicolumn{2}{c}{32.023/3.183} 
& \multicolumn{2}{c}{39.901/1.428} 
& \multicolumn{2}{c}{38.045/1.345} 
& \multicolumn{2}{c}{38.293/1.479}  \\
\bottomrule

\end{tabular}
}
\caption{CPU time and standard deviation (Average Time / Standard Deviation) for the empirical coverage and average interval width reported in Table \ref{tab:simulation}.}
\label{tab:comp_costs}
\end{table}

Table \ref{tab:comp_costs} presents the corresponding average CPU time (in seconds) and its standard deviation under 100 iterations (for non-bootstrap simulations) and 1,000 iterations (for bootstrap simulations) across the same simulation scenarios and modelling frameworks as Table \ref{tab:simulation}. We can observe that the split CP exhibits near-instantaneous computation times across all scenarios, averaging between 0.001 and 0.018 seconds regardless of sample size or model complexity. This efficiency stems from its working regime: a single model fit on the training subset followed by straightforward quantile calculations on the calibration set. Standard deviations remain negligible ($\leq 0.007$ seconds), which indicates consistent runtime stability. This makes the split method ideal for real-time applications or large-scale iterative analyses where computational resources are constrained. In contrast, the full CP exhibits substantially higher computational costs due to its need to refit models for each candidate point. Additionally, the full CP presents significant runtime variability (with a standard deviation up to 0.119 seconds). This volatility seems to stem from algorithmic sensitivity to random covariate effects and optimization convergence in dispersion parameters. Despite these computational demands, the full CP's more efficient intervals remain an advantage. Practitioners must therefore weigh the critical trade-off: split conformal for computational efficiency when chasing for the speed versus full conformal for statistical precision when resources permit. We also observe that generally, as $\phi$ increases (or $\sigma$ decreases), the computational time for the full CP decreases, which may be because the model is easier to fit under smaller variability. The computational costs for other scenarios are presented in Table \ref{tab:extra_costs}.

\begin{table}[ht]
\centering
\resizebox{\textwidth}{!}{%
\begin{tabular}{ll*{12}{c}}
\toprule
\multirow{2}{*}{\textbf{Scenario}} & \multirow{2}{*}{\textbf{Sample Size}} 
& \multicolumn{4}{c}{$\phi=2$, $\sigma=1.5$} 
& \multicolumn{4}{c}{$\phi=5$, $\sigma=0.9$} 
& \multicolumn{4}{c}{$\phi=20$, $\sigma=0.45$} \\
\cmidrule(lr){3-6} \cmidrule(lr){7-10} \cmidrule(lr){11-14}
& & S1-Split & S1-Full & S3-Split & S3-Full 
  & S1-Split & S1-Full & S3-Split & S3-Full 
  & S1-Split & S1-Full & S3-Split & S3-Full \\
\midrule

\multirow{4}{*}{\textbf{M1}} 
& $n=50$   & 0.001/0.001 & 0.033/0.006 & 0.001/0.001 & 0.034/0.010 & 0.001/0.001 & 0.030/0.005 & 0.001/0.001 & 0.031/0.005 & 0.001/0.001 & 0.029/0.005 & 0.001/0.001 & 0.028/0.004 \\
& $n=100$  & 0.001/0.001 & 0.034/0.007 & 0.001/0.001 & 0.034/0.005 & 0.001/0.001 & 0.032/0.007 & 0.001/0.001 & 0.032/0.004 & 0.001/0.001 & 0.030/0.006 & 0.001/0.001 & 0.030/0.005 \\
& $n=500$  & 0.001/0.001 & 0.038/0.009 & 0.001/0.001 & 0.038/0.006 & 0.001/0.001 & 0.035/0.006 & 0.001/0.001 & 0.035/0.006 & 0.001/0.001 & 0.032/0.008 & 0.001/0.001 & 0.033/0.006 \\
& $n=1000$ & 0.001/0.001 & 0.043/0.009 & 0.001/0.001 & 0.043/0.006 & 0.001/0.001 & 0.039/0.006 & 0.001/0.001 & 0.042/0.022 & 0.001/0.001 & 0.036/0.006 & 0.001/0.001 & 0.036/0.006 \\
\midrule

\multirow{4}{*}{\textbf{M2}} 
& $n=50$   & 0.008/0.003 & 0.272/0.052 & 0.008/0.003 & 0.275/0.049 & 0.008/0.003 & 0.259/0.047 & 0.008/0.004 & 0.270/0.058 & 0.008/0.003 & 0.242/0.042 & 0.008/0.003 & 0.253/0.044 \\
& $n=100$  & 0.007/0.003 & 0.259/0.033 & 0.007/0.003 & 0.262/0.042 & 0.008/0.003 & 0.241/0.031 & 0.007/0.003 & 0.255/0.049 & 0.007/0.003 & 0.227/0.028 & 0.007/0.003 & 0.244/0.041 \\
& $n=500$  & 0.008/0.003 & 0.324/0.034 & 0.007/0.003 & 0.343/0.052 & 0.008/0.003 & 0.311/0.032 & 0.007/0.003 & 0.337/0.048 & 0.008/0.003 & 0.299/0.025 & 0.007/0.003 & 0.310/0.041 \\
& $n=1000$ & 0.009/0.006 & 0.429/0.050 & 0.008/0.003 & 0.457/0.055 & 0.008/0.003 & 0.414/0.043 & 0.008/0.003 & 0.453/0.057 & 0.009/0.004 & 0.408/0.050 & 0.008/0.003 & 0.428/0.051 \\
\midrule

\multirow{4}{*}{\textbf{M3(P)}} 
& $n=50$   & 0.006/0.008 & 0.113/0.018 & 0.005/0.002 & 0.114/0.025 & 0.005/0.002 & 0.112/0.015 & 0.004/0.001 & 0.111/0.023 & 0.007/0.006 & 0.121/0.015 & 0.005/0.002 & 0.116/0.027 \\
& $n=100$  & 0.006/0.006 & 0.141/0.024 & 0.006/0.004 & 0.142/0.024 & 0.005/0.002 & 0.136/0.028 & 0.006/0.004 & 0.138/0.011 & 0.005/0.002 & 0.125/0.027 & 0.005/0.002 & 0.118/0.015 \\
& $n=500$  & 0.010/0.004 & 0.363/0.023 & 0.010/0.003 & 0.370/0.041 & 0.010/0.003 & 0.326/0.057 & 0.011/0.003 & 0.368/0.047 & 0.007/0.003 & 0.236/0.011 & 0.007/0.002 & 0.243/0.035 \\
& $n=1000$ & 0.015/0.004 & 0.667/0.038 & 0.016/0.004 & 0.675/0.034 & 0.015/0.005 & 0.665/0.101 & 0.016/0.004 & 0.670/0.089 & 0.010/0.003 & 0.409/0.035 & 0.011/0.004 & 0.423/0.031 \\
\midrule

\multirow{4}{*}{\textbf{M3(Q)}} 
& $n=50$   & 0.006/0.005 & 0.119/0.029 & 0.005/0.002 & 0.112/0.010 & 0.005/0.004 & 0.111/0.026 & 0.004/0.002 & 0.107/0.009 & 0.007/0.003 & 0.120/0.018 & 0.005/0.002 & 0.113/0.013 \\
& $n=100$  & 0.005/0.002 & 0.138/0.009 & 0.005/0.002 & 0.143/0.026 & 0.005/0.002 & 0.132/0.016 & 0.005/0.002 & 0.136/0.012 & 0.006/0.002 & 0.123/0.016 & 0.005/0.002 & 0.118/0.025 \\
& $n=500$  & 0.010/0.003 & 0.366/0.028 & 0.010/0.003 & 0.368/0.019 & 0.009/0.003 & 0.324/0.060 & 0.010/0.003 & 0.366/0.047 & 0.008/0.002 & 0.241/0.028 & 0.007/0.002 & 0.242/0.010 \\
& $n=1000$ & 0.015/0.004 & 0.656/0.038 & 0.016/0.005 & 0.680/0.045 & 0.013/0.004 & 0.663/0.101 & 0.015/0.005 & 0.672/0.092 & 0.011/0.004 & 0.417/0.033 & 0.011/0.003 & 0.419/0.032 \\
\midrule

\multirow{4}{*}{\textbf{M4(P)}} 
& $n=50$   & 0.006/0.002 & 0.130/0.037 & 0.006/0.002 & 0.132/0.017 & 0.007/0.006 & 0.130/0.014 & 0.006/0.002 & 0.127/0.012 & 0.007/0.003 & 0.126/0.028 & 0.007/0.004 & 0.123/0.023 \\
& $n=100$  & 0.007/0.002 & 0.162/0.019 & 0.007/0.003 & 0.177/0.030 & 0.006/0.002 & 0.151/0.023 & 0.007/0.004 & 0.143/0.009 & 0.006/0.003 & 0.142/0.011 & 0.006/0.002 & 0.140/0.009 \\
& $n=500$  & 0.010/0.003 & 0.401/0.054 & 0.012/0.008 & 0.507/0.061 & 0.011/0.005 & 0.383/0.030 & 0.011/0.004 & 0.352/0.023 & 0.010/0.004 & 0.326/0.024 & 0.010/0.003 & 0.327/0.025 \\
& $n=1000$ & 0.017/0.005 & 0.612/0.038 & 0.023/0.007 & 0.636/0.069 & 0.017/0.004 & 0.702/0.036 & 0.016/0.004 & 0.661/0.031 & 0.014/0.003 & 0.580/0.048 & 0.016/0.007 & 0.569/0.042 \\
\midrule

\multirow{4}{*}{\textbf{M4(Q)}} 
& $n=50$   & 0.006/0.002 & 0.130/0.024 & 0.006/0.002 & 0.137/0.016 & 0.006/0.002 & 0.128/0.011 & 0.006/0.002 & 0.130/0.013 & 0.007/0.003 & 0.128/0.027 & 0.006/0.002 & 0.126/0.012 \\
& $n=100$  & 0.006/0.002 & 0.169/0.036 & 0.006/0.002 & 0.183/0.031 & 0.007/0.002 & 0.151/0.022 & 0.006/0.002 & 0.149/0.023 & 0.006/0.003 & 0.143/0.008 & 0.006/0.002 & 0.147/0.022 \\
& $n=500$  & 0.010/0.004 & 0.398/0.043 & 0.010/0.003 & 0.512/0.072 & 0.011/0.004 & 0.378/0.027 & 0.011/0.003 & 0.363/0.024 & 0.011/0.005 & 0.331/0.031 & 0.010/0.003 & 0.335/0.023 \\
& $n=1000$ & 0.018/0.007 & 0.618/0.035 & 0.020/0.004 & 0.646/0.072 & 0.016/0.003 & 0.694/0.036 & 0.016/0.004 & 0.673/0.035 & 0.014/0.003 & 0.574/0.034 & 0.014/0.004 & 0.568/0.034 \\
\bottomrule
\end{tabular}
}
\caption{CPU time and standard deviation (Average Time / Standard Deviation) for the empirical coverage and average interval width reported in Table \ref{tab:extra_simulation}.}
\label{tab:extra_costs}
\end{table}

\section{Empirical Evidence for the Application}
\label{app:application}

\begin{figure}[ht]
    \centering
    \begin{subfigure}[t]{0.48\textwidth}
        \centering
        \includegraphics[width=\linewidth]{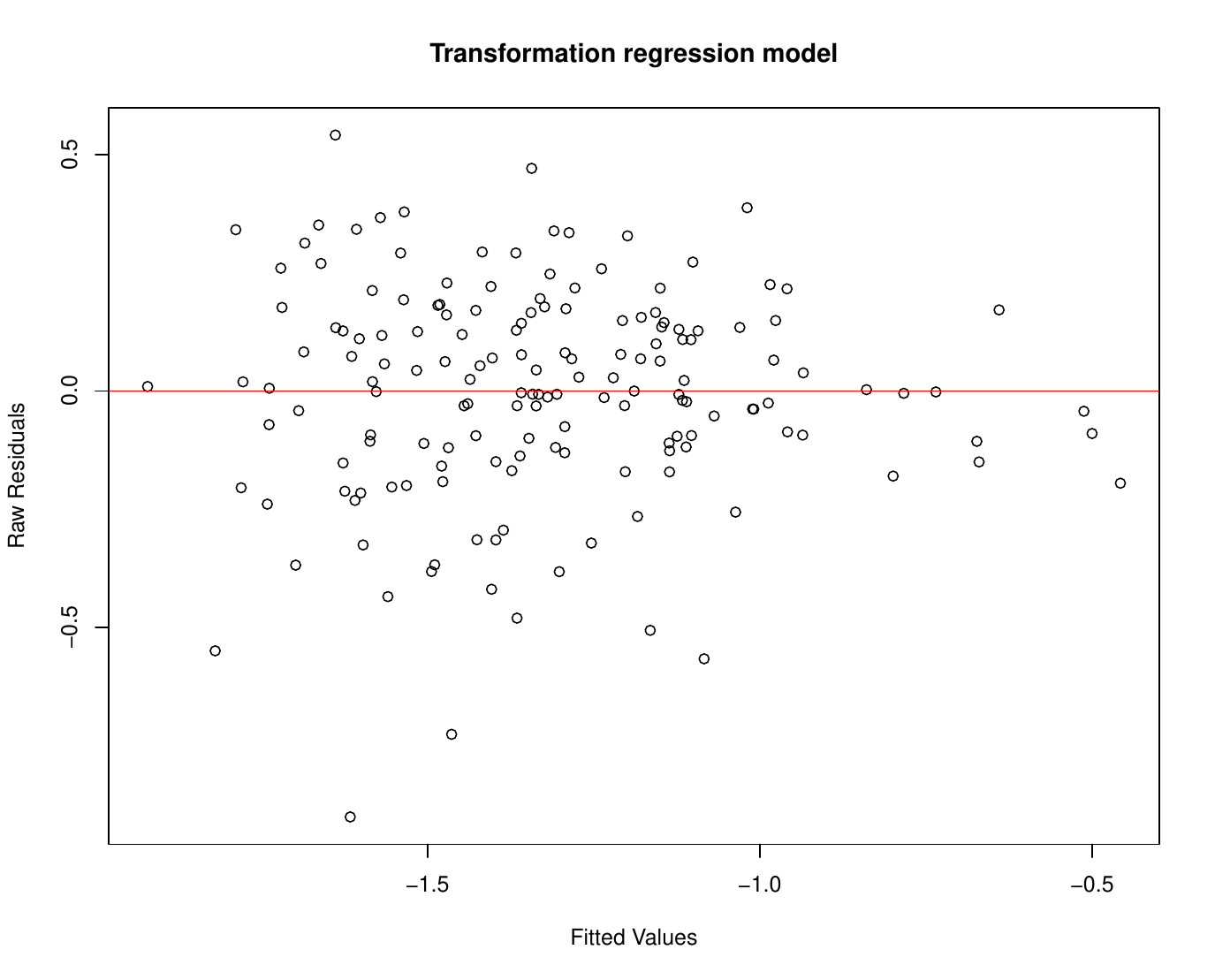}
        \caption{Residuals vs fitted values for the logit-normal model.}
        \label{fig:resid_trans}
    \end{subfigure}
    \hfill
    \begin{subfigure}[t]{0.48\textwidth}
        \centering
        \includegraphics[width=\linewidth]{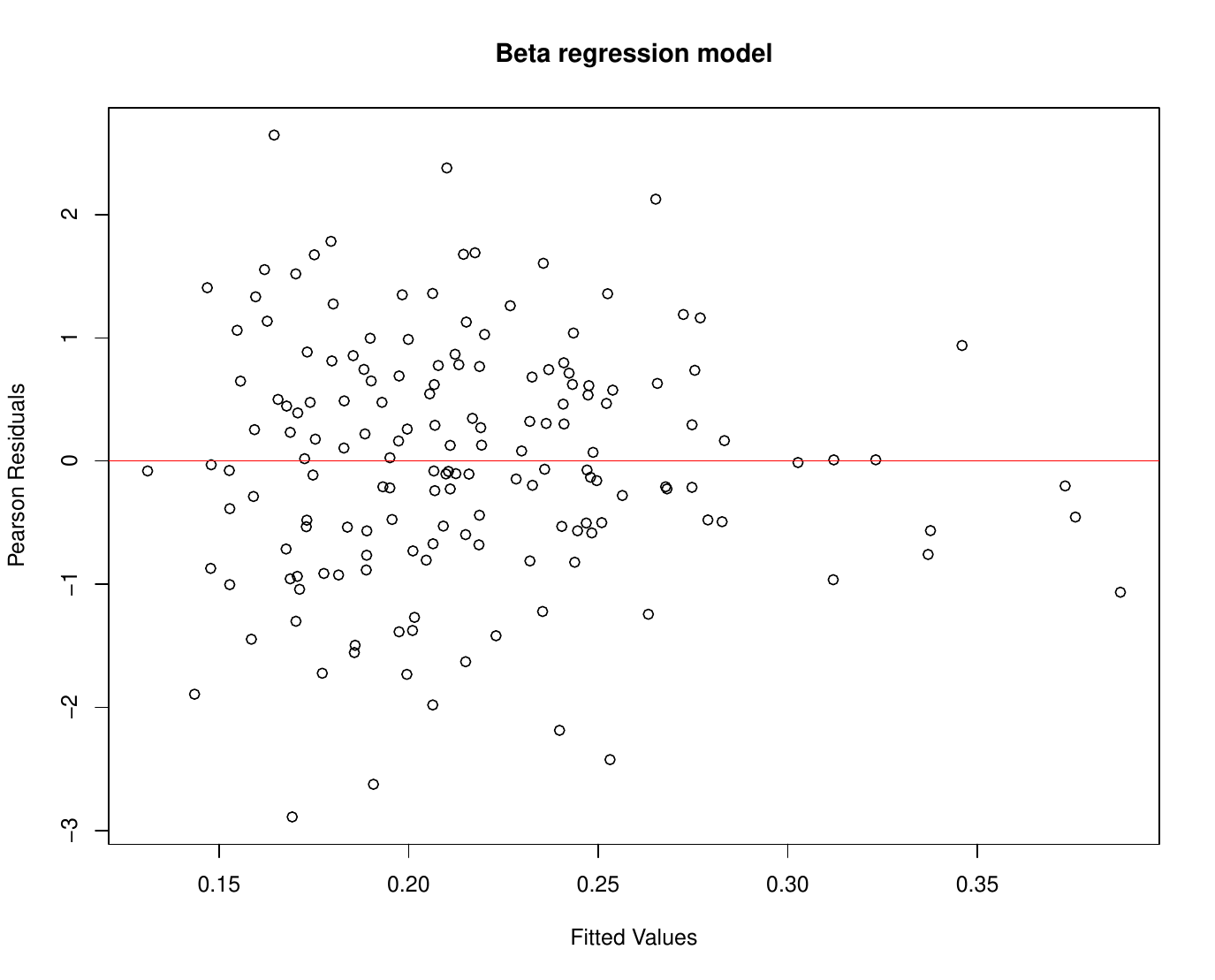}
        \caption{Pearson residuals vs fitted values for the beta model.}
        \label{fig:resid_beta}
    \end{subfigure}
    \caption{Residual plots for the two model frameworks.}
    \label{fig:residuals}
\end{figure}

As further empirical evidence motivating the introduction of heteroscedasticity into CP models, we present the residual-versus-fitted plots using the training set plus the calibration set for both the logit-normal model and the beta regression model in Figure \ref{fig:residuals}. These plots reveal the non-constant variance patterns that can support the introduction of heteroscedasticity in the construction of conformal and bootstrap prediction intervals. For the raw residual plot in the logit-normal regression model, the residuals are approximately centered around zero, but there is a funnel-shaped pattern suggesting that the spread of residuals may vary with the level of the fitted values. For the Pearson residual in the beta regression model, the pattern is more pronounced here, with increased residual dispersion observed at higher fitted values. These diagnostic plots therefore provide visual signal for the need of heteroscedastic conformal prediction models in our analysis.

\vskip 0.2in

\bibliography{sample}

\end{document}